\documentclass[final]{cvpr}

\usepackage{times}
\usepackage{epsfig}
\usepackage{graphicx}
\usepackage{amsmath}
\usepackage{amssymb}
\usepackage{amsfonts}       
\usepackage{nicefrac}       
\usepackage{microtype}      
\usepackage{mathtools}
\usepackage{amsmath}
\usepackage{bm}
\usepackage{bbm}
\usepackage{stackengine}
\usepackage{multirow}
\usepackage{verbatim}
\usepackage{color}
\usepackage{float}
\usepackage{enumitem}
\usepackage{booktabs}
\usepackage{tabulary,multirow,overpic,xcolor}
\usepackage[caption=false]{subfig}
\usepackage{pifont}
\usepackage{epstopdf}
\usepackage[nosort]{cite}

\definecolor{Gray}{gray}{0.5}
\definecolor{GrayBG}{gray}{0.95}
\usepackage[british,UKenglish,USenglish,english,american]{babel}

\newcommand{\tblref}[1]{Table~\ref{#1}}
\newcommand{\sref}[1]{Sec.~\ref{#1}}

\usepackage{soul}
\definecolor{carmine}{rgb}{0.59, 0.0, 0.09}

\newcommand{\app}{\raise.17ex\hbox{$\scriptstyle\sim$}}

\def\x{$\times$}
\newcolumntype{x}[1]{>{\centering\arraybackslash}p{#1pt}}

\newlength\savewidth\newcommand\shline{\noalign{\global\savewidth\arrayrulewidth
		\global\arrayrulewidth 1pt}\hline\noalign{\global\arrayrulewidth\savewidth}}
\newcommand{\tablestyle}[2]{\setlength{\tabcolsep}{#1}\renewcommand{\arraystretch}{#2}\centering\footnotesize}

\makeatletter\renewcommand\paragraph{\@startsection{paragraph}{4}{\z@}
	{.5em \@plus1ex \@minus.2ex}{-.5em}{\normalfont\normalsize\bfseries}}\makeatother

\definecolor{citecolor}{RGB}{34,139,34}
\definecolor{lightred}{RGB}{241,140,142}
\definecolor{citecolor2}{HTML}{0071bc}

\usepackage[pagebackref=true,breaklinks=true,letterpaper=true,colorlinks,
citecolor=citecolor2,bookmarks=false]{hyperref}

\renewcommand{\omega}{\alpha}
\renewcommand{\phi}{\beta}

\def\cvprPaperID{***} 
\def\confYear{CVPR 2021}
\begin{document}
	
	\title{ A Large-Scale Study on Unsupervised Spatiotemporal Representation Learning\\ }

	\author{
		Christoph Feichtenhofer \qquad
		Haoqi Fan \qquad
		Bo Xiong \qquad
		Ross Girshick  \qquad
		Kaiming He \vspace{.8em}\\
		Facebook AI Research (FAIR)
	}
	
	\maketitle
	
	\definecolor{fastcolor}{RGB}{121,178,128}
	\definecolor{slowcolor}{RGB}{165,170,243}
	\newcommand{\fastcolor}[1]{\textcolor{fastcolor}{\textbf{#1}}}
	\newcommand{\fastcolorC}[1]{\textcolor{orange}{#1}}
	\newcommand{\slowcolor}[1]{\textcolor{slowcolor}{#1}}
	
	\newcommand{\slow}{\slowcolor{Slow }}
	\newcommand{\fast}{\fastcolor{Fast }}
	
	\definecolor{predictioncolor}{RGB}{0,255,0}
	\definecolor{labelcolor}{RGB}{255,0,0}
	\newcommand{\predictioncolor}[1]{\textcolor{predictioncolor}{#1}}
	\newcommand{\labelcolor}[1]{\textcolor{labelcolor}{#1}}
	
	\newcommand{\pred}{\predictioncolor{\textbf{Predictions}: }}
	\newcommand{\gt}{\labelcolor{\textbf{Labels}: }}

	\definecolor{demphcolor}{RGB}{144,144,144}
	\newcommand{\demph}[1]{\textcolor{demphcolor}{#1}}
	
	
	\begin{abstract}
		We present a large-scale study on unsupervised spatiotemporal representation learning from videos. With a unified perspective on four recent image-based frameworks, we study a simple objective that can easily generalize all these methods to space-time. Our objective encourages temporally-persistent features in the same video, and in spite of its simplicity, it works surprisingly well across: (i) different unsupervised frameworks, (ii) pre-training datasets, (iii) downstream datasets, and (iv) backbone architectures. We draw a series of intriguing observations from this study, e.g., we discover that encouraging long-spanned persistency can be effective even if the timespan is 60 seconds. In addition to state-of-the-art results in multiple benchmarks, we report a few promising cases in which unsupervised pre-training can outperform its supervised counterpart. Code is made available at  \url{https://github.com/facebookresearch/SlowFast}.
	\end{abstract}
	\section{Introduction}
	\label{sec:introduction}
	
A series of recent methods on unsupervised representation learning from images~\cite{He20, Chen20, Grill2020, Caron20} are based on maximizing a similarity objective for different views of the same image under data augmentations \cite{Dosovitskiy16a,Wu18}. In addition to the \mbox{\emph{artificial}} augmentations on images, videos can provide \mbox{\emph{natural}} augmentations of visual content under various changing factors, such as motion, deformation, occlusion, and illumination. This work aims to generalize these image-based methods \cite{He20, Chen20, Grill2020, Caron20} into space-time.

	\begin{figure}[t]
		\centering
		\includegraphics[width=0.9\linewidth]{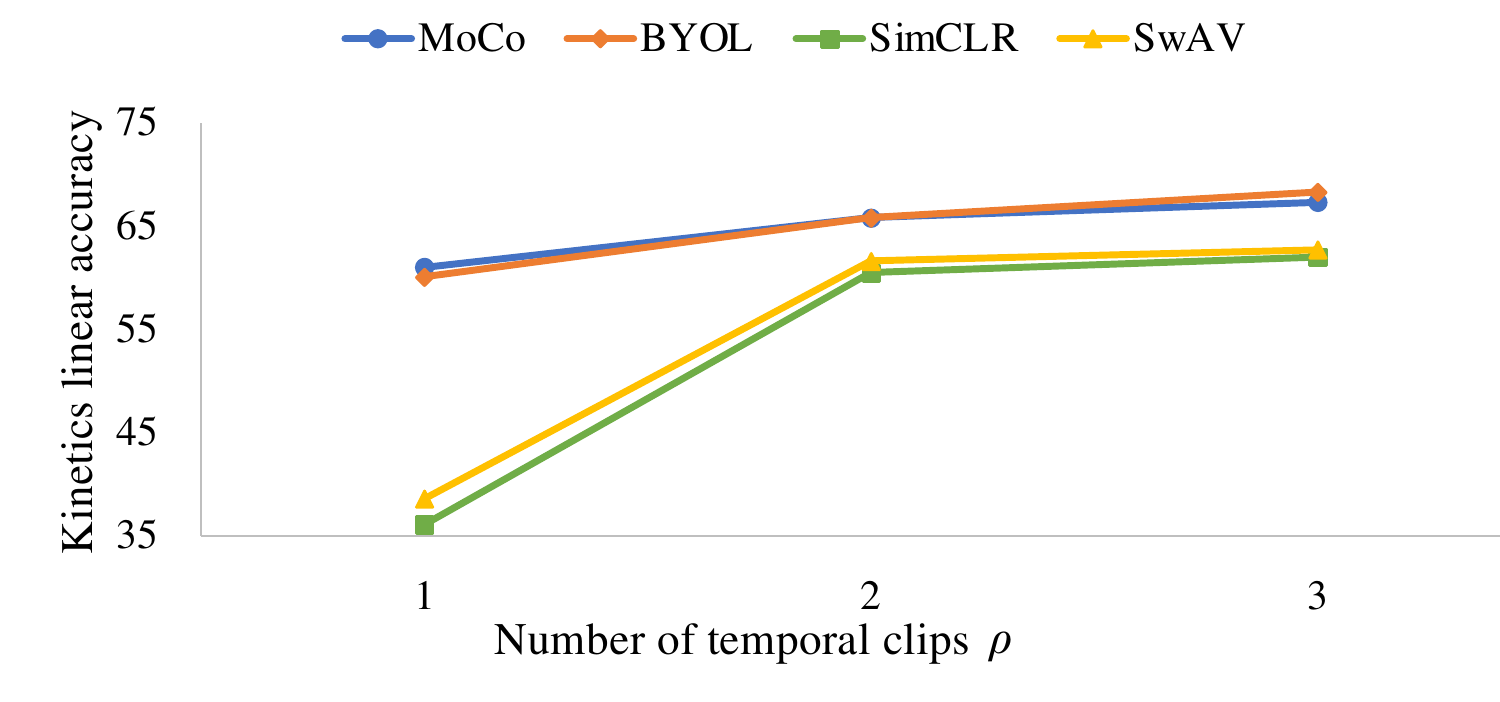} 
		\vspace{-0.5pt}
		\includegraphics[width=0.9\linewidth]{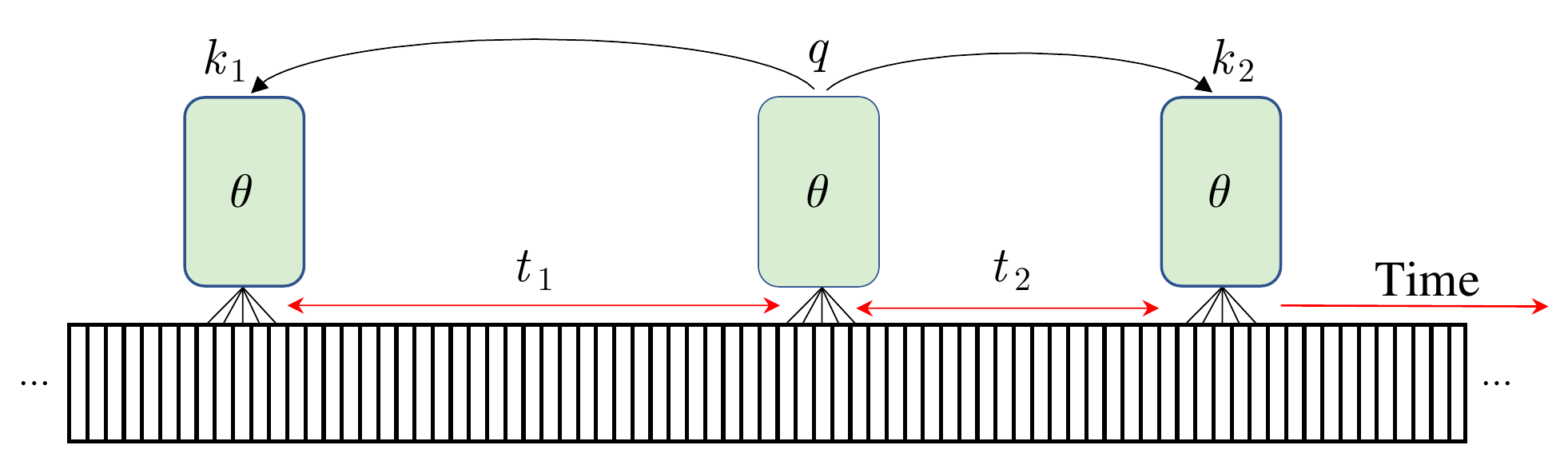}
		\caption{
Learning to maximize the similarity between different temporal clips of the same video encourages feature persistency over time. A query clip ($q$) is matched to multiple key clips ($k_1, k_2, \ldots $) that are temporally shifted. This method can be incorporated into several unsupervised learning frameworks (MoCo\cite{He20}, SimCLR~\cite{Chen20}, BYOL~\cite{Grill2020}, SwAV~\cite{Caron20}).
The figure on the top shows that increasing the number ($\rho$) of temporal clips improves representation quality for all these frameworks. 
		}
		\vspace{-1.5em}
		\label{fig:teaserConcept}
	\end{figure}

We study a simple objective that can be easily incorporated into these image-based methods. Our hypothesis is that the visual content is often \emph{temporally-persistent} along a timespan in the video. This persistency may involve an action (\eg, a person dancing), an object (\eg, an individual person, who transitions from running to walking), and a scene (\eg, a room with people moving), covering short to long spans, with different levels of visual invariance (action, object, scene). Our objective simply encourages \emph{the visual representations in different clips of the same video to be similar}. We empirically find that this objective works well across different unsupervised frameworks (MoCo\cite{He20}, SimCLR~\cite{Chen20}, BYOL~\cite{Grill2020}, SwAV~\cite{Caron20}), either \emph{with} or \emph{without} using dissimilar (negative) samples.

Our objective is a natural generalization of \emph{crops} in images \cite{Dosovitskiy16a,Wu18} to \emph{clips} in videos. This allows us to make use of the recent unsupervised learning frameworks with minimal modifications. 
We aim to learn a high-level representation of the categorical semantics present in a video by enforcing persistency of the representation over space-time. We investigate factors such as the effective timespan, $t$, between positives, and number of temporal clips, $\rho$, to find that longer timespans (up to a minute) and multiple samples  are beneficial for downstream performance (\figref{fig:teaserConcept}).

Our unsupervised training is performed on large-scale data, including Kinetics \cite{Kay17} (240k videos) and three versions of \emph{million-scale} Instagram sets. In addition to standard linear probing, we  evaluate representation quality on multiple classification and detection downstream datasets, \eg, \mbox{Charades~\cite{Sigurdsson2016}, Something-Something \cite{ssv2}, and AVA \cite{Gu2018}.} 

Our results suggest that unsupervised pre-training can achieve competitive performance in videos, and it can surpass the supervised pre-training counterparts in a few cases. Finally, our study also reveals room for improvement along multiple directions.
	
In summary, our large-scale study involves the following five aspects:

(i) Four unsupervised learning frameworks (MoCo\cite{He20}, SimCLR~\cite{Chen20}, BYOL~\cite{Grill2020}, SwAV~\cite{Caron20}) viewed from a unified perspective, and incorporated with a simple temporal persistency objective;

(ii) Three pre-training datasets, including the relatively well-controlled Kinetics~\cite{Kay17} and the relatively ``in-the-wild" Instagram sets at million-scale;

(iii) Six downstream datasets/tasks for evaluating representation quality;

(iv) Ablation experiments on different factors, such as temporal samples, contrastive objective, momentum encoders, training duration, backbones, data augmentation, curated \vs uncurated, trimmed \vs untrimmed, \etc; and

(v) State-of-the-art results of unsupervised video representation learning on established benchmarks, UCF-101~\cite{Soomro12}, HMDB51~\cite{Kuehne11} and Kinetics-400~\cite{Kay17} . 	
	
	\begin{figure*}[t]\centering
		\hfill
		\subfloat[SimCLR]{\includegraphics[width=0.22\linewidth]{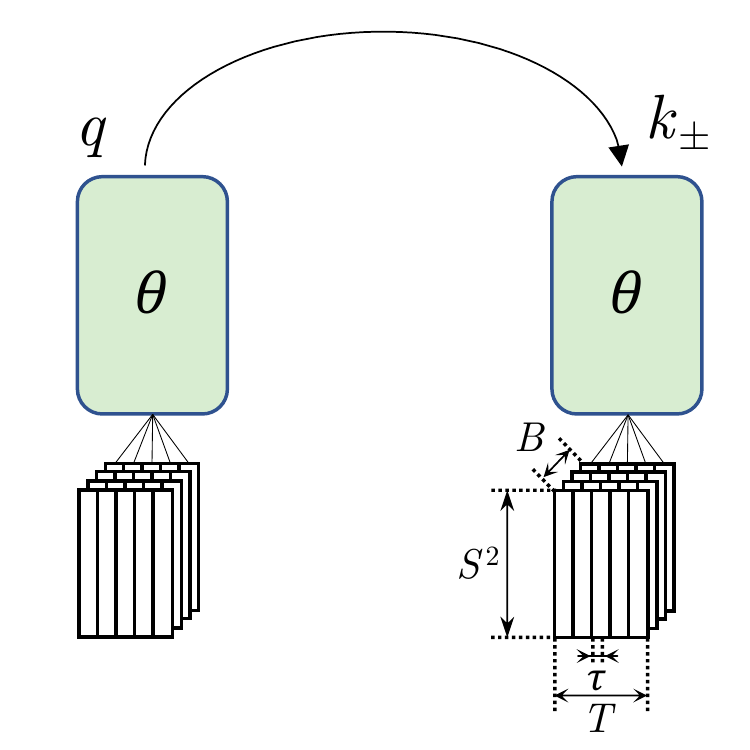}\label{fig:methods:simclr}}
		\hfill
		\subfloat[MoCo]{\includegraphics[width=0.22\linewidth]{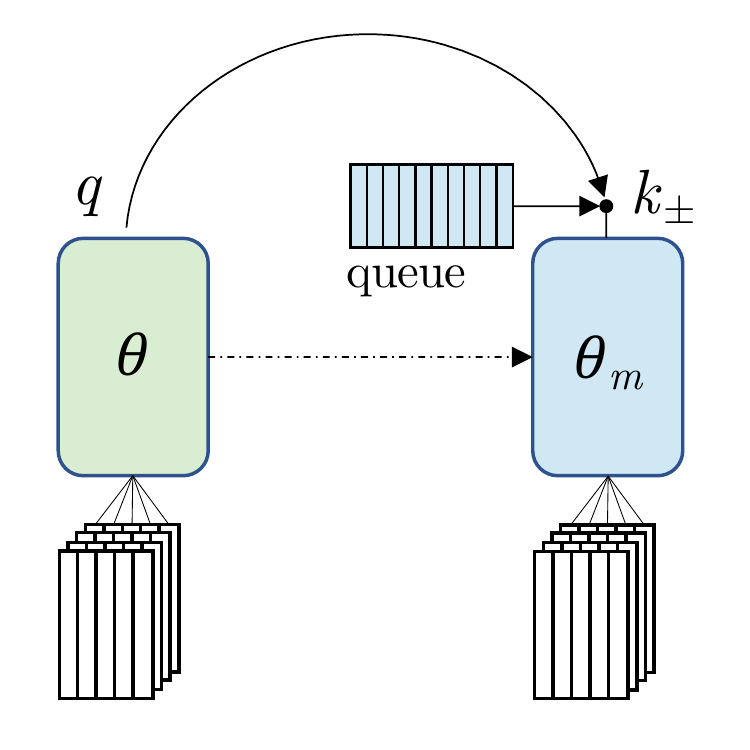} \label{fig:methods:moco}}
		\hfill
		\subfloat[BYOL]{\includegraphics[width=0.22\linewidth]{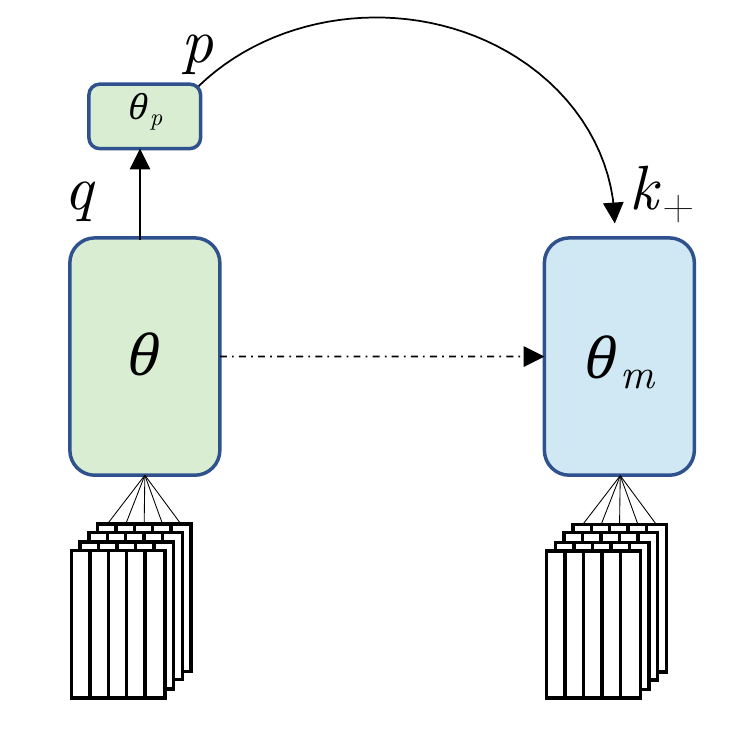} \label{fig:methods:byol}}
		\hfill
		\subfloat[SwAV]{\includegraphics[width=0.22\linewidth]{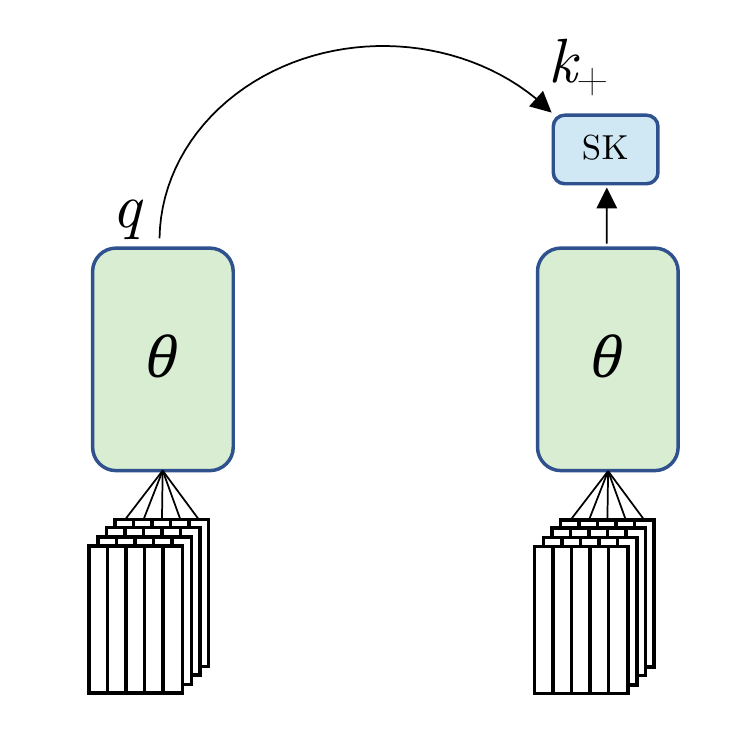}\label{fig:methods:swav}}
		\hfill
		\vspace{5pt}
		\caption{\textbf{Conceptual comparison of four unsupervised learning mechanisms applied to video}.	The inputs consist of  $\rho{=}$2 clips from $B$ videos. Each clip is a stack of $T$ frames with temporal stride $\tau$ and spatial resolution $S^2$. Each method trains encoder weights $\theta$ by computing a positive loss component \wrt to the other clips of the same video.  SimCLR \protect\subref{fig:methods:simclr} and MoCo \protect\subref{fig:methods:moco} use a contrastive loss with negatives coming from different videos in the batch or a a queue. respectively. MoCo \protect\subref{fig:methods:moco} and BYOL \protect\subref{fig:methods:byol} use extra momentum encoders with weights $\theta_m$ being moving averages of the trained $\theta$. SwAV \protect\subref{fig:methods:swav}  uses a Sinkhorn-Knop (SK) transform to generate the positive targets. 	\label{fig:methods}
		}	
	\end{figure*}

	\section{Related Work}
	\label{sec:related_work}
	
	\paragraph{Unsupervised learning in images} has been actively researched recently with approaches focusing on various pretext tasks related to color- or patch-based processing \cite{Pathak16, Zhang16color,Doersch15,Noroozi16}, instance discrimination with contrastive objectives \cite{Dosovitskiy16a,Wu18,Oord18,Henaff19,Hjelm19,Ji19,He20,Zhuang19,Chen20,Tian20} and ones that focus on positive pairs~\cite{Caron18,Caron20,Grill2020}.

	\paragraph{Unsupervised learning in videos} has followed a similar trajectory with earlier methods focusing on predictive tasks based on motion, color and spatiotemporal ordering~\cite{goroshin2015unsupervised,Isola15,Agrawal15,Jayaraman15,Srivastava15,Wang15,Misra16,Vondrick16b,Mathieu16,Lotter17,Fernando17,Lee17,Wang17,Pathak17,Gan18,Kim19,Xu19vcop,Diba19,Wang19,purushwalkam2020demystifying,jenni2020video}, and contrastive objectives with visual~\cite{Sermanet18,Sun19,Han19,Li20a,Gordon20,yang2020video} and audio-visual input~\cite{Owens16,Arandjelovic17,Arandjelovic18,Korbar18,Alwassel19,Patrick20,Piergiovanni20}.
	
	Several recent ones~\cite{Gordon20,Han20,Alwassel19,Patrick20,alayrac2020self,Qian20,yang2020video,morgado2020audio} relate to image-based approaches~\cite{He20,Caron18,Chen20,Wu18}.
	With some of them using additional modalities of optical-flow~\cite{Tian20,Han20}, audio~\cite{Alwassel19,Patrick20,alayrac2020self,morgado2020audio} and text~\cite{Sun19,alayrac2020self} to transfer supervision from one modality to another.  
	
	In relation to these previous efforts, our work studies purely visual unsupervised learning from video and tries to compare the meta-methodologies on common ground.
	
	\paragraph{Evaluation protocols and backbones} in most image-based approaches have converged to ResNet-50~\cite{He16} encoders with ImageNet linear-classification protocol, and several smaller downstream tasks~\cite{He20, Chen20, Grill2020, Caron20} for evaluation. ~ In video understanding research, the field has not yet converged and is using different backbones with focus on finetuning performance on two relatively small datasets~\cite{Soomro12,Kuehne11}. We investigate this aspect by looking at different encoders and 6 different downstream benchmarks for evaluation.
	
	\section{Approach} \label{sec:approach}
	The objective of this work is to study several recent unsupervised representation learning methodologies to train a spatiotemporal encoder $f_\theta$, exploring implementation details and comparing them on a common ground to measure their efficacy in video understanding. We focus on two contrastive approaches using positive and negative samples: SimCLR \cite{Chen20} and MoCo \cite{He20}, as well as two approaches that solely rely on  positives, BYOL \cite{Grill2020} and  SwAV \cite{Caron20}  (\sref{sec:contrastive}).
	
	These approaches were originally presented for learning image representations, and they all share the objective of learning invariant features across different views (crops/augmentations) of the spatial image input. In this paper, this idea is extended to the temporal domain. 
	Our core idea is to learn an encoder $f_\theta$ that produces embeddings  which are persistent in space-time, over multiple ($\rho$) temporally distant clips of the same video. This is related to Slow Feature Analysis \cite{Wiskott02} where the objective is to minimize the representations' temporal derivative over the input. The general idea of learning temporally persistent features is not new and has been proposed in the past with similar motivation \eg, \cite{Becker97, Mobahi09, goroshin2015unsupervised}.
	
	\subsection{Persistent temporal feature learning}
	Our framework takes different augmented clips $x$ \ of an unlabeled video and passes them through an encoder $f_\theta$ with weights $\theta$ to obtain corresponding embeddings $q = f_\theta(x)$. The encoder is spatiotemporal ConvNet, by default a \mbox{ResNet-50} (R-50) \cite{He16}, Slow-only pathway of SlowFast Networks \cite{Feichtenhofer19}, which is a 3D ResNet-50 \cite{He16} without temporal pooling in convolutional feature maps, followed by an MLP projection head, that produces and output of dimension $d$.  
	
	The input clips are  stacks of RGB frames of size \mbox{$3\times T\times S^2$} for  temporal \x~spatial dimensions, which are sampled with temporal stride $\tau$ , \ie, the encoder processes only one out of $\tau$ frames of the raw video. Therefore, $T \times \tau$ define the timespan and resolution of the encoder. 
	
	Given a minibatch of $B$ videos, our framework creates a set of $\rho B$ positive examples by sampling $\rho$ clips from the videos. The learning methodologies studied in this section maximize  similarity of a ``query'' sample $q$ with a set of positive ``key'' samples $\{k^{+}\}$ that are encoded  versions of different clips of the same video as $q$ is computed from. 
	\figref{fig:teaserConcept} illustrates an example where $\rho{=}$3 clips are used.
	
	The next section describes how the contrastive and non-contrastive unsupervised representation learning methodologies are exemplified.
	\subsection{Unsupervised learning frameworks} \label{sec:contrastive}
	Contrastive learning maximizes the similarity of a sample $q$ with positive ones  $\{k^{+}\}$ and minimizes similarity to negative ones $\{k^{-}\}$. The contrastive approaches in this paper use the  \mbox{InfoNCE} \cite{Oord18} objective,
	
	\begin{equation}\label{eq:infonce} 	\small
		\mathcal{L}_{q} = -\log{\frac{ {\sum_{k \in \{k^+\}}} \exp\left({\mathrm{sim}(q, k) / \alpha}\right)}{ {\sum_{k \in \{k^+, k^-\}}} {\exp\left({\mathrm{sim}(q, k) / \alpha}\right)}   }},
	\end{equation}
	with $\alpha$ being a temperature hyper-parameter for scaling and $\{k^{+}\}$ are embedded clips of the same video as $q$.
	All the embeddings are $\ell_2$  normalized and dot product (cosine) similarity is used to compare them $\mathrm{sim}(q, k) = q^\top k / \lVert q\rVert \lVert k\rVert$.

	\paragraph{SimCLR} \cite{Chen20} (\figref{fig:methods:simclr}) uses the embeddings of clips of other videos in the minibatch as negatives $\{k^{-}\}$.
	
	\paragraph{MoCo} \cite{He20} (\figref{fig:methods:moco}) is a method that uses an explicit momentum encoder which parameters, $\theta_m$, are a moving average \mbox{$\theta_m \leftarrow m \theta_k + (1 - m) \theta$} with $m$ a momentum parameter. In eq.~\eqref{eq:infonce} MoCo uses this encoder to compute the positive embeddings $\{k^{+}\}$ from clips of the same video as $q$, and negative embeddings $\{k^{-}\}$ are taken from a queue that stores embeddings of clips from previous iterations. There is no backpropagation into the momentum-encoder weights $\theta_m$.
	
	\paragraph{BYOL} \cite{Grill2020} (\figref{fig:methods:byol}) can be viewed as a form of MoCo that does not use negative samples, but an extra predictor MLP with weights $\theta_p$, which is stacked on top of $f_\theta$'s MLP head. For a sample $q = f_{\theta_p}(f_\theta(x))$, BYOL minimizes negative cosine similarity,
	\begin{equation}\label{eq:byol} 	\small
				\mathcal{L}_{q} = - {\sum_{k \in \{k^+\}}}  \mathrm{sim}(q, k) = - {\sum_{k \in \{k^+\}}}  q^\top k^+ / \lVert q\rVert \lVert  k\rVert,
	\end{equation}
	with $\{k^{+}\ {=} f_{\theta_m}(x^+) \}$ being embedded clips $x^+$ from the same video as $q$, encoded with momentum weights $\theta_m$.

	\paragraph{SwAV}  \cite{Caron20} (\figref{fig:methods:swav}) can be viewed as a form of SimCLR that does not use negative samples. SwAV first performs a linear mapping of the positive embeddings $q,k^+$ to learned prototypes $\tilde{q},\tilde{k}^+$ and then transforms the targets with an extra Sinkhorn-Knopp ($SK$) step. Then the SwAV loss is
	\begin{equation}
		\mathcal{L}_{q} =   D_\text{KL}(  \tilde{q} \| SK(\tilde{k}^+) ),
		\label{eq:swav}
	\end{equation}
	where $D_\text{KL}$ is the The Kullback-Leibler divergence and gradients are not back-propagated through the $SK$ operation. 
	
		 Compared to SimCLR and MoCo, in BYOL and SwAV,  $q$ and  $k$  are not typical ``query'' and ``key'' samples (but rather ``source'' and ``target'' samples); however, for consistency we use $q$, $k$ terminology in notation for all methods.
	 
	 \paragraph{Implementation specifics.} 
	 
	We implement the methods with a \textit{symmetric} loss, as in original SimCLR, BYOL and SwAV, where every input clip is used to produce a loss (and gradient) signal. For each of the $\rho\geq2$ clips, we compute $q$, while all \textit{other} $\rho{-}1$ clips of the same video are used as $\{k^{+}\}$ to evaluate sub-loss $\mathcal{L}_{q}$	
	and the symmetric loss is the average over all $\rho$  sub-losses.
 Thus, for MoCo and BYOL, every input clip is processed by both encoders.

	For MoCo and BYOL, our symmetric loss is aggregated \textit{sequentially} which implies that memory consumption for $\rho>2$ equals to a single clips' forward and backward pass, since these methods do not backpropagate through the momentum encoder. For  SimCLR and SwAV the overall loss is evaluated in \textit{parallel} across all clips and therefore memory consumption grows linearly with the number of clips used. 
	
	All details on implementation and pre-training are in \S\ref{sec:training}. 
	
	\section{Experiments} \label{sec:experiments}
	
	\begin{table}[h!]
		\vspace{-10pt}
		\centering
		\small
		\tablestyle{2pt}{1.05}
		\begin{tabular}{l|x{24}|rrrrrr}
			\multicolumn{1}{c|}{data} & \#videos &$t_\text{median}$ &$t_\text{mean}$ & $t_\text{std}$ & $t_\text{min}$ & $t_\text{max}$ \\
			\shline
			Kinetics-400 (K400) \cite{Kay17} & 240K & 10.0 & 9.3 & 1.7 & 1.0& 10.0 \\
			IG-Curated \cite{Ghadiyaram2019}  & 1M & 18.9 & 26.3 & 19.8 & 1.5  &  60.0  \\
			IG-Uncurated & 1M & 29.4 &  35.3 & 38.4 & 0.5  &  600.0  \\
			IG-Uncurated-Short & 1M & 13.0 &  13.1 & 1.6 & 10.0  &  15.9  \\
			
		\end{tabular}
		\vspace{.3em}
		\caption{\textbf{Pre-training data statistics with timings in seconds}.
		}
		\label{tab:data_statistics} 
		\vspace{-10pt}
	\end{table}
	
	\paragraph{Datasets.} Unless otherwise noted, we perform unsupervised pre-training on Kinetics-400 \cite{Kay17} (K400)  with $\app$240k training videos in 400 human action categories. 
	
	To study learning from ``\textit{in-the-wild}'' videos from the web, we pre-train the methods on Instagram videos: \mbox{IG-Curated}~\cite{Ghadiyaram2019}, a dataset with hashtags similar to K400 classes; IG-Uncurated which has videos taken randomly from Instagram; and IG-Uncurated-Short which is similar, but has constrained duration. Each dataset has 1M videos.

	\tblref{tab:data_statistics} shows dataset statistics of all datasets used for unsupervised pre-training. Most of Kinetics videos are of 10 seconds in duration. IG-Curated is a dataset with Instagram videos that have an average duration $t_\text{mean}$ of 26.3 seconds and a standard deviation $t_\text{std}$ of 29.8 seconds. The maximum duration $t_\text{max}$ is 60s. IG-Uncurated contains videos taken randomly from Instagram, with larger deviation in length and maximum duration of 10 minutes (600s). IG-Uncurated-Short is a dataset consisting of random Instagram videos that have a duration between 10 and 16 seconds, to study the effect of a fixed duration and the assumption that short videos may hold more useful information for pre-training.
	
	\paragraph{Evaluation protocols.} 
	For evaluation we use two protocols. 
	
	The first one is common to evaluate unsupervised image representations~\cite{He20, Chen20}. It validates the \textit{linear classifier} performance based on frozen encoder features that are taken from the global average pooling layer. We report top-1 classification accuracy (\%) on the K400 validation set. 
	
	The second protocol reports \textit{finetuning} accuracy on the first split of the UCF101 dataset~\cite{Soomro12} which contains 13k videos in 101 human action classes; this is a common procedure used to evaluate unsupervised video representations.
	Finally, we also report \textit{finetuning} accuracy on AVA~\cite{Gu2018},  Charades~\cite{Sigurdsson2016}, Something-Something~\cite{ssv2} and HMDB51 \cite{Kuehne11}.
	
	\paragraph{Architecture.} 
	By default, we use  a R-50 \cite{He16} following the Slow pathway in \cite{Feichtenhofer19} with clips of $T{=}$8 frames sampled with stride $\tau{=}$8 from 64 raw-frames of video. The supervised performance for training 200, 400, 800 epochs on K400 is  74.7\%, 74.3\% and 72.7\%, respectively, and does not improve for training longer due to overfitting. 

	\paragraph{Implementation details.} We follow default settings in video classification~\cite{Feichtenhofer19}. Specifics on the approaches, their training and evaluation and the impact of implementation on performance are provided in \S\ref{sec:impl_details} and \S\ref{sec:results_impl}, respectively.
	
	\begin{table}[t] 
		\vspace{-10pt}
		\centering
		\tablestyle{1pt}{1.05} 
		\begin{tabular}{x{10}|x{26}x{26}|x{26}x{26}|x{26}x{26}|x{26}x{26}}
			&  \multicolumn{2}{c|}{ \textbf{MoCo} } &  \multicolumn{2}{c|}{ \textbf{BYOL} }  &  \multicolumn{2}{c|}{ \textbf{SimCLR}} &  \multicolumn{2}{c}{ \textbf{SwAV}} \\
			\multicolumn{1}{c|}{$\rho$}  & K400 & UCF101 & K400 &UCF101 & K400 &UCF101 & K400 &UCF101  \\
			\shline
			1 &  61.0 & 90.8 & 60.6 & 91.2 & 36.1 & 84.2 & 38.6 & 74.7  \\ 
			2 & 65.8 & 91.0  & 65.8 & 92.7 & 60.5 & 88.9 & 61.6	& 87.3 \\ 
			3 & 67.3 & 92.8 & 68.3 & 93.8 & 62.0 & 87.9 & 62.7 & 89.4   \\
			4 & 67.8 & 93.5 & \textbf{68.9} & \textbf{93.8} &  \multicolumn{4}{c}{ out of memory}   \\
		\end{tabular}
		\vspace{.3em}
		\caption{\textbf{Number of  temporal clips $\rho$}. Data: \textbf{K400}, 200 epochs. Learning temporally persistent features ($\rho\geq2$) is effective. }
		\label{tab:crops_vs_clips}
	\end{table}
	
	\subsection{Persistent temporal learning} \label{sec:persistent_ablations}
	Here, we investigate the impact of learning spatiotemporal \vs only spatial persistent features.  \tblref{tab:crops_vs_clips} shows the accuracy of the four methods when trained for 200 epochs on K400, and evaluated on K400 (linear) and UCF101 (finetuned), \ie our default setting. 
	
	\paragraph{Temporal augmentation.}
	
	The first row in \tblref{tab:crops_vs_clips}, $\rho{=}$1, uses two spatial crops at the same temporal instance, while the $\rho{=}$2 row uses clips at different temporal locations as positives; therefore, learns persistent features in time. This difference has a large impact on performance, especially for SimCLR (60.5 $\rightarrow$ 36.1) and SwAV (61.6 $\rightarrow$ 38.6) performance degrades significantly when sampling positives from the same temporal instance ($\rho{=}$1). 
	\paragraph{More clips are beneficial.}
	The remaining rows in \tblref{tab:crops_vs_clips}  show that accuracy is further increasing with the number of temporal samples per video, \eg at $\rho{=}$4 the best accuracy is achieved with \textbf{BYOL} at \textbf{68.9}\% K400 and \textbf{93.8}\% UCF101.
	
	\paragraph{Negatives do not help but momentum encoders do.}	
	When comparing the methods in \tblref{tab:crops_vs_clips}, we see that: 
	
	(i) There is no clear performance difference between contrastive/non-contrastive methods. This indicates that learning space-time persistence within a video is key for the methods, but learning in-persistence across videos is not. 
	
	(ii) There is a clear difference of $\app$4\% on K400 between methods that employ momentum encoders (MoCo, BYOL), \vs these that do not (SimCLR, SwAV). 
	
	\begin{table}[t] 
		\centering
		\tablestyle{1pt}{1.05} 
		\begin{tabular}{x{14}|x{26}x{26}|x{26}x{26}|x{26}x{26}|x{26}x{26}}
			&  \multicolumn{2}{c|}{ \textbf{MoCo} } &  \multicolumn{2}{c|}{ \textbf{BYOL} }  &  \multicolumn{2}{c|}{ \textbf{SimCLR}} &  \multicolumn{2}{c}{ \textbf{SwAV}} \\
			\multicolumn{1}{c|}{ep}  & K400 &UCF101 & K400 &UCF101 & K400 &UCF101 & K400 &UCF101  \\
			\shline
			50 & 52.6 & 84.6  &  30.2 &	78.5  &  45.7 & 79.7  & 55.9 & 81.4    \\ 
			100  & 60.5 & 89.5  & 47.6	& 88.6  & 57.3	& 85.6 & 59.4 & 85.5  \\
			200 &   65.8 & 91.0 &  {65.8} & 92.7 &  60.5 & 88.9 & 61.6 & 87.3 \\
			400 & 67.4 & 92.5 & 66.9 & 92.8 & 62.0 & 87.9 & 62.9 & 88.3  \\ 
			800 & \textbf{67.}4 & \textbf{93.2} & 66.2 & 93.6  & 61.8 & 88.4 & 63.2 & 89.5 \\ 
		\end{tabular}
		\vspace{.3em}
		\caption{\textbf{Training duration in epochs (ep)}:  Dataset: \textbf{K400},  $\rho{=}$2. Training longer brings consistent gains for all methods up to 400 epochs and saturates for K400 but not for UCF101 at 800ep. SwAV is the strongest performer for short training (50ep). }
		\label{tab:epochs}
	\end{table}

	Increasing the number of clips per training iteration increases training cost, so it is reasonable to compare it to training more epochs. \tblref{tab:epochs} is studying the base case  $\rho{=}$2 for various number of epochs (ep).
	
	Overall, the results show that there is a clear gain for training longer which has been also observed in image-related tasks \cite{Chen20,He20,Grill2020,Caron20}. 
	BYOL performs the worst when training short durations. This might be related to hyper-parameter settings which we do not adjust for this experiment (the original implementation \cite{Grill2020}  uses different hyper-parameters for different number of training epochs). 
	
	\subsection{Timespan between positives}
		All experiments with $\rho{\geq}$2 so far were using global temporal sampling of positives, which means that the clips can be sampled at unconstrained temporal locations from the input video. This might be counter-productive because if there is a long duration that has passed between a pair of positive clips they might no longer share the same semantic context for learning high-level features corresponding in time. 
		
	\begin{table}[h!]\centering
		\captionsetup[subfloat]{captionskip=2pt}
		\captionsetup[subffloat]{justification=centering}
		\subfloat[ Dataset: \textbf{K400}, 200 epochs training.
		\label{tab:delta_max_k400}]{
			\tablestyle{5pt}{1.05} 
			\begin{tabular}{ccccccccc}
				$t_\text{max}$ in seconds & 0 & 2 & 3  & 4 & 5 & 8 & 10 \\
				\shline
				K400 acc in \%  & 60.6 & 65.2 & 65.7 & 65.8 & 65.8 &  65.6 & 65.8 \\
		\end{tabular}}\\ \vspace{-5pt}
		\subfloat[  Dataset: \textbf{IG-Curated-1M}, 50 epochs training.
		\label{tab:delta_max_ig_kin}]{
			\tablestyle{6pt}{1.05} 
			\begin{tabular}{ccccccccc}
				$t_\text{max}$ in seconds &4 &  8 & 16 & 32 & 60 \\
				\shline
				K400 acc in \% &   62.7   & 63.1 & 63.1  & 63.9  & 64.1 \\
		\end{tabular}}\\ \vspace{-5pt}
		\subfloat[  Dataset: \textbf{IG-Uncurated-1M}, 50 epochs training.
		\label{tab:delta_max_ig_rnd}]{
			\tablestyle{6pt}{1.05} 
			\begin{tabular}{ccccccccc}
				$t_\text{max}$ in seconds &12s &  24 & 36 & 48 & 600\\
				\shline
				K400 acc in \%  &   59.3   & 59.2 & 59.9  & 59.6  & 58.9 \\
		\end{tabular}}
		\vspace{0.5em}
		\caption{\textbf{Maximum frame distance for positives}. Method: \textbf{BYOL}, $\rho=$~2. Training is surprisingly robust with increasing accuracy for increased distance between samples. Accuracy only (mildly) degrades when sampling positives that are more than ~36 seconds apart when using uncurated (random) videos. 
		}
		\label{tab:delta_max}
	\end{table}
	
	This experiment is concerned with the maximum distance between the positive training samples. We use BYOL pre-training on K400, IG-Curated-1M and IG-Uncurated-1M and report 400 linear readout accuracy in \tblref{tab:delta_max}.
	
	\tblref{tab:delta_max_k400} shows performance for increasing the maximum temporal distance between positives in \textbf{K400} pre-training. It can be seen that using positives from the same time ($t_\text{max}{=}$0) degrades perforance b $\app$5\% but other than that performance is relatively robust up to global sampling of positive clips from the whole video ($t_\text{max}{=}$10s). This is interesting as it seems that a long-temporal correspondence objectives does not hurt performance (but also does not boost it).
	
	\tblref{tab:delta_max_ig_kin} shows performance for increasing the temporal distance between positive samples on \textbf{IG-Curated-1M}. This dataset has a maximum duration of 60 seconds; statistics are in \tblref{tab:data_statistics}. \tblref{tab:delta_max_ig_kin} reveals that increasing the maximum duration between positive pairs is beneficial for performance and unrestricted sampling of positives is the best with 64.1\% top-1 accuracy for evaluation on K400. This is especially interesting, as it shows that even longer videos benefit from  global sampling. There is \textit{no benefit from restricting the time window of positives}, which can be interpreted as the objective of learning extremely-slow features~\cite{Wiskott02} that do not change over 60 seconds of video. Long-temporal-distance samples might also increase robustness of the model by providing ``\textit{hard-positive}'' samples for learning. Note that here the videos are still sampled according to hashtags related to K400 classes~\cite{Ghadiyaram2019}; therefore, the conjecture might be biased. 
	
	Finally, we are looking at the \textbf{IG-Uncurated-1M} dataset which consists of a random sampling of 1M videos from Instagram. These videos can be between 0.5s and 10 minutes of duration. Most of the videos however are much shorter than 10 minutes, with a mean duration of 35.3 seconds and a standard deviation of 38.4 seconds (\tblref{tab:data_statistics}). For this data, \tblref{tab:delta_max_ig_rnd} shows the results of progressively increasing the maximum timespan between positive samples. It can be observed that increasing the maximum distance between positives up to 36 seconds is beneficial and beyond that performance decreases, but only slightly, even when performing global sampling of positives (the default).
		
	\subsection{Backbone architectures} \label{sef:backbones}
		So far all experiments were using a R-50, 8\x8 Slow pathway \cite{He16,Feichtenhofer19} as backbone. The next set of ablations studies different architectures for the spatiotemporal encoder.
		
	\begin{table}[h!] 
		\centering
		\tablestyle{3.2pt}{1.1} 
		\begin{tabular}{l|c|ccc|c|cc}
			& &\multicolumn{3}{c|}{training}   &  \multicolumn{1}{c|}{sup.} & \multicolumn{2}{c}{\textbf{MoCo} ($\rho{=}$2)}    \\
			\multicolumn{1}{c|}{ backbone } & ${T \times \tau}$  & FLOPs & Param & s/iter & K400 & K400 &UCF101 \\
			\shline
			R-50   &8\x8  & 41.7G & 31.8M & 1.6s & 74.7  & 65.8 &  91.0  \\ \hline 
			R-18   &8\x8  &  20.0G  & 20.2M & 1.2s & 68.9  & 56.2 & 87.1 \\   
			R-101   &8\x8  & 93.3G & 51.4M & 2.1s &  75.8  & 67.7 & 92.4 \\ \hline 
			
			R-50   & 16\x4  & 83.5G & 31.8M & 2.5s  & 76.1  & 67.6 & 93.3 \\ 
			R-50   & 32\x2  & 167.0G & 31.8M &4.6s & 76.3 & 67.8 & 94.2 \\ \hline 
			
			R2+1D-18 & 32\x2  & 48.5G &15.4M & 4.0s & 71.7 & 57.2 &  93.7 \\  
			
			S3D-G  & 32\x2  & 36.0G & 9.1M & 4.1s & 74.7  & 63.2  &  94.5 \\  
		\end{tabular}
		\caption{\textbf{Backbone comparison.} The ResNet~\cite{He16} backbone (Slow pathway~\cite{Feichtenhofer19}) is used with different depth (R-18, R-50, R-101), input frames $T$ and stride $\tau$. R2+1D \cite{Tran18} and S3D-G \cite{Xie18s3d} are commonly used backbones for unsupervised video representation learning with downstream evaluation on UCF101. }
		\label{tab:architectures}
	\end{table}
	
	\tblref{tab:architectures} compares different backbones for usage with MoCo in our default setting ($\rho{=}$2, 200 epoch pre-training on K400). From left to right, the table shows the input duration $T$, sampling-rate $\tau$, FLOPs (at 224$^2$ spatial resolution) and parameters of these backbones, as well as the average duration for training one iteration of the MoCo algorithm (measured on a single machine with 8 V100 GPUs in \texttt{PySlowFast} \cite{fan2020pyslowfast} and \texttt{torchvision} decoder), the supervised performance on K400 and UCF101 (finetuned from K400), as well as the downstream performance for K400 linear evaluation and UCF101 finetuning. 
	
	The first observation in \tblref{tab:architectures} is that for the Slow architecture~\cite{Feichtenhofer19}, using shallower (R-18) or deeper (R-101) networks can influence supervised and downstream performance in a sizable manner, with MoCo, K400 evaluation benefiting from more parameters. Doubling the input frame-rate (8\x8 $\rightarrow$ 16\x4) boosts accuracy on UCF101.
	
	The second observation is that R2+1D~\cite{Tran18} has a large gap on Kinetics (71.7\% supervised \vs 57.2\% unsupervised), while being remarkably strong on UCF101 (93.7\%). This gap is also observed for S3D-G~\cite{Xie18s3d}. The reason for this might be that UCF101 is a small dataset which is easy to overfit and can benefit from fewer parameters. 

	\begin{table}[h!] 
		\centering
		\captionsetup[subfloat]{captionskip=2pt}
		\captionsetup[subffloat]{justification=centering}

		\subfloat[Training on \textbf{IG-Curated-1M}.
		\label{tab:epochs-IG-kin}]{
			\hspace{-10pt}
			\tablestyle{1pt}{1.05} 
			\begin{tabular}{x{14}|x{26}x{26}|x{26}x{26}|x{26}x{26}|x{26}x{26}}
				&  \multicolumn{2}{c|}{ \textbf{MoCo} } &  \multicolumn{2}{c|}{ \textbf{BYOL} }  &  \multicolumn{2}{c|}{ \textbf{SimCLR}} &  \multicolumn{2}{c}{ \textbf{SwAV}} \\
				\multicolumn{1}{c|}{ep}  & K400 &UCF101 & K400 &UCF101 & K400 &UCF101 & K400 &UCF101  \\
				\shline
				50 &  64.8	& 91.1 & 64.1	& 93.5 & 55.5 & 86.4 & 61.0 &	89.0 \\ 
				200  &  69.0 & 93.4 & 60.2 & 92.7 & 56.9 & 86.6 & 64.3 &	91.2  \\
		\end{tabular}} \vspace{-10pt}
		\subfloat[Training on \textbf{IG-Uncurated-1M}. 	\label{tab:epochs-IG-R} ]{
			\tablestyle{1pt}{1.05} 
			\hspace{-10pt}
			\begin{tabular}{x{14}|x{26}x{26}|x{26}x{26}|x{26}x{26}|x{26}x{26}}
				&  \multicolumn{2}{c|}{ \textbf{MoCo} } &  \multicolumn{2}{c|}{ \textbf{BYOL} }  &  \multicolumn{2}{c|}{ \textbf{SimCLR}} &  \multicolumn{2}{c}{ \textbf{SwAV}} \\
				\multicolumn{1}{c|}{ep}  & K400 &UCF101 & K400 &UCF101 & K400 &UCF101 & K400 &UCF101  \\
				\shline
				50 & 61.8 &	90.9 & 58.9	& 90.1 & 52.1 & 85.1 &  56.0 & 86.7 \\ 
				200  & 65.4	& 91.9 &  57.9 & 91.6  & 51.9 &	85.3 & 58.8&87.8\\
		\end{tabular}} \vspace{-10pt}
		\subfloat[Training on \textbf{IG-Uncurated-Short-1M}. 	\label{tab:epochs-IG-RS} ]{
			\tablestyle{1pt}{1.05} 
			\hspace{-10pt}
			\begin{tabular}{x{14}|x{26}x{26}|x{26}x{26}|x{26}x{26}|x{26}x{26}}
				&  \multicolumn{2}{c|}{ \textbf{MoCo} } &  \multicolumn{2}{c|}{ \textbf{BYOL} }  &  \multicolumn{2}{c|}{ \textbf{SimCLR}} &  \multicolumn{2}{c}{ \textbf{SwAV}} \\
				\multicolumn{1}{c|}{ep}  & K400 &UCF101 & K400 &UCF101 & K400 &UCF101 & K400 &UCF101  \\ 
				\shline
				50 & 61.0 &	89.6  & 62.3 &	91.4 & 53.61 &	86.4  & 55.0	& 86.2 \\ 
				200 & 64.5	& 91.0  & 57.0	& 90.9 & 55.97	& 86.9 & 58.4	&87.2  \\
		\end{tabular}}
		\caption{Training on curated \protect\subref{tab:epochs-IG-kin}, uncurated \protect\subref{tab:epochs-IG-R} and short duration video \protect\subref{tab:epochs-IG-RS} data from the web. Longer training degrades performance for BYOL, possibly due to suboptimal hyper-parameters.  $\rho{=}$2. }
		\label{tab:epochs-IG}
	\end{table}
	
		\subsection{Uncurated data and  video duration} \label{sec:ig_scale}
		
	In \tblref{tab:epochs-IG} we show the performance of all four methodologies on 
	{IG-Curated-1M} \protect\subref{tab:epochs-IG-kin}, {IG-Uncurated-1M} \protect\subref{tab:epochs-IG-R} and {IG-Uncurated-Short-1M} \protect\subref{tab:epochs-IG-RS} for pre-training with 50 and 200 epochs.
	We make the following observations: 
	
	(i) Among the methods MoCo performs the best with \eg 69.0\% \vs  second-best 64.3\% of SwAV on curated data \protect\subref{tab:epochs-IG-kin}. 
	
	(ii) MoCo and SwAV scale the best for training longer, gaining roughly 3-4\% for 200ep \vs 50ep.
	
	(iii) On uncurated data, MoCo and SwAV perform $\app$1\% better on the unconstrained duration videos in \protect\tblref{tab:epochs-IG-R}. 
	
	(iv)\ BYOL and SimCLR show better performance on IG-Uncurated-Short (10-16s videos) in \tblref{tab:epochs-IG-RS}, seemingly benefiting from shorter videos, but there is no clear benefit from either longer or shorter duration among all methods.
	
	(v) BYOL degrades performance for training longer which might be due to the requirement of different hyper-parameters for different schedules (as noted in \sref{sec:persistent_ablations}). 
	
	We will return to this point in  \S\ref{sec:extra_IG}, where we show that increasing clips-size $\rho$ can overcome this issue in BYOL, along with further studies on the trade-off against training more epochs, and dataset scale. 
		
	\subsection{Data augmentations} \label{sec:results_augmentations}
	
\begin{figure}[t!]
		\centering
		\vspace{-2em}
		\includegraphics[width=1\linewidth]{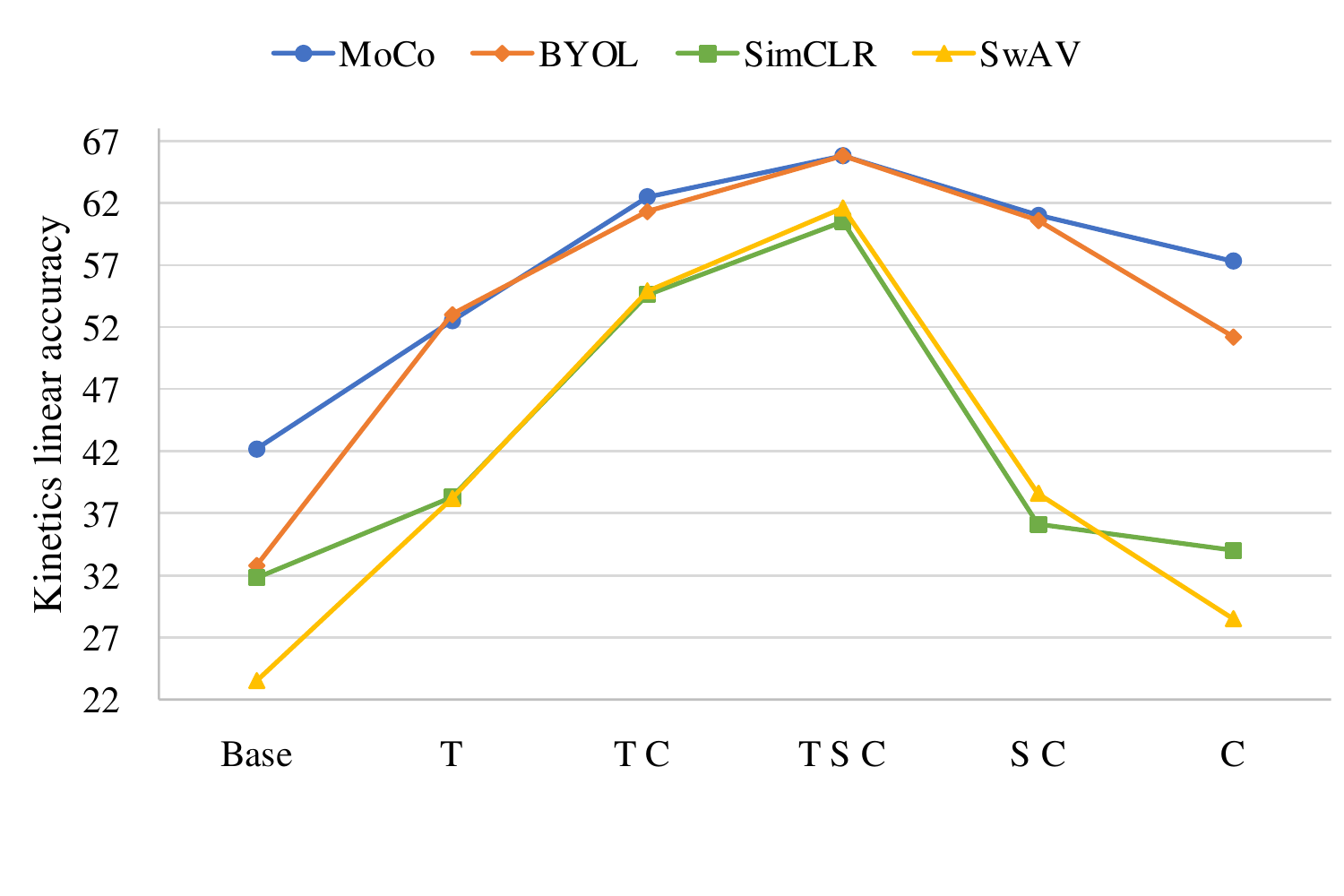}
		\vspace{-3em}
		\caption{\textbf{Ablating augmentations.}  We explore temporal (T), spatial (S),  and color (C) augmentations to learn persistent features.
		}
		\label{fig:aug_removal}
	\end{figure}

	\paragraph{Importance of augmentations.} 	Augmentations can have a major impact on visual unsupervised feature learning~\cite{Chen20,Chen20moco}. In  \figref{fig:aug_removal}, we ablate spatial cropping (S), temporal clipping (T) and radiometric color (C) augmentations from the four unsupervised learning methods (\eg ``T S C'' are the baselines using all augmentations and removing ``S C'' equals $\rho{=}$1 in \tblref{tab:crops_vs_clips}). We make three main observations: 
	
	(i) Among the methods, MoCo and BYOL perform most robust for using fewer augmentations; their advantage over SimCLR and SwAV might be related to the momentum encoder which can provide extra augmentation in training. 
	
	(ii) When minimizing the augmentations by  resizing the shorter size of the video to the input size of 224 and only cropping along the long side of the video (Base in \figref{fig:aug_removal}), MoCo still provides 42.2\% K400 linear accuracy, over \mbox{BYOLs'} 32.4\%, showing an advantage of the contrastive loss in a weak augmentation scenario.
	
	(iii) Among the augmentations, learning temporal (T) persistency, has the largest impact on performance, except for MoCo which benefits more from color (C) (incl.~grayscale) augmentations. Especially SimCLR and SwAV show significant drops in performance when removing T, \ie when extracting positive clips from the same instance in time. 
	
	In the remainder of this section, we explore using stronger augmentations than the default ones in previous experiments. We perform the ablations with MoCo in the basic setting of $\rho=$~2, 200 epochs K400 pre-training.

	\begin{table}[h!] 
		\centering
		\tablestyle{3pt}{1.05} 
		\tablestyle{6.4pt}{1.0}	
		\begin{tabular}{cccccc}
			
			\multicolumn{1}{c}{color } & grayscale & temporal & fps & \multicolumn{2}{c}{accuracy}  \\
			strength & probability &  difference & jitter & K400 & UCF101  \\
			\shline
			0.5    & 0.2  &  &  & 65.8 &  91.0  \\ \hline
			0.75    & 0.2  &  &   & \textbf{66.0} &  \textbf{92.1}  \\
			1.0   & 0.2  &  &   & 65.8 &  91.2  \\ \hline 
			0.5    & 0.4  &  &  & 65.5 &  91.0  \\ \hline
			0.5    & 0.2  & \checkmark &  & \textbf{66.2} &  91.3 \\
			0.5    & 0.2  &  & \checkmark & 65.6 & 91.5  \\
		\end{tabular}
		\vspace{.3em}
		\caption{\textbf{Radiometric augmentation}. Method: \textbf{MoCo}, 200 epochs, $\rho=$~2. Dataset: \textbf{K400}. Stronger color augmentation in K400 pre-training can especially benefit UCF101 (+1.3\%).} \vspace{-5pt}
		\label{tab:color_aug}
	\end{table}
		
	\paragraph{Stronger color augmentation.}
	In  \tblref{tab:color_aug} color strength of 0.5 indicates the default one for MoCo \cite{Chen20moco}, 0.75 and 1.0 increase the strength of randomly jittering brightness, contrast, saturation and hue proportionally. 
	
	\tblref{tab:color_aug}  shows that increasing it to 0.75 can improve K400/UCF101 accuracy. Increasing the random grayscale probability from 0.2 to 0.4 does not provide an improvement on either of the datasets. However, using a temporal-difference augmentation which randomly (with probability 0.2) first converts the frames to grayscale and then subtracts them across time, can increase K400 accuracy by 0.4\%. Finally, using frame-rate jittering of $\pm$50\% of the original frame-rate does not improve K400 but UCF101 slightly.
	
	\begin{table}[h!] 
		\vspace{-10pt}
		\centering
		\tablestyle{3pt}{1.05} 
		\tablestyle{6.4pt}{1.0}	
		\begin{tabular}{cccccc}
			\multicolumn{2}{c}{area } &  aspect  & \multicolumn{2}{c}{accuracy}  \\
			$[$ min, & max$]$  &  ratio &  K400 & UCF101  \\
			\shline
			\multicolumn{2}{c}{default \cite{Simonyan15, He16, Feichtenhofer19}}  &  & 65.8 &  91.0 \\ \hline
			$[$0.49, & 0.76$]$  & & 64.8 & 91.7 \\ 
			$[$0.49, & 0.76$]$  & \checkmark& 65.4 & 91.7 \\ 
			$[$0.20, & 0.76$]$  & \checkmark& \textbf{66.8} & \textbf{91.8} \\ 
			$[$0.20, & 0.50$]$  & \checkmark& 66.3 & 91.8 \\ 
			$[$0.20, & 1.00$]$  &\checkmark& 66.6 & 91.7 \\ 
			$[$0.08, & 0.50$]$  & \checkmark& 64.3 & 91.6 \\ 
			$[$0.08, & 1.00$]$  &\checkmark& 65.3 & 91.2 \\ 
		\end{tabular}
		\vspace{.3em}
		\caption{\textbf{Cropping augmentation}. Method: \textbf{MoCo}, 200 epochs, $\rho=$~2. Dataset: \textbf{K400}. Stronger cropping and aspect ratio augmentation can be beneficial by +1.0\% (K400) and 0.7\% UCF101. } \vspace{-10pt}
		\label{tab:crop_aug}
	\end{table}
	\paragraph{Spatial cropping.}
	Our default implementation uses VGG-style \cite{Simonyan15, He16} cropping that randomly resizes the \textit{shorter spatial side} of a video between [256, 320] pixels and takes a random 224$^\text{2}$ crop extended over time to extract a clip \cite{Feichtenhofer19}. 
	
	Since unsupervised learning might benefit from more aggressive cropping, we explore Inception-style \cite{Szegedy15} cropping with aspect ratio augmentation that is commonly used in unsupervised learning from images~\cite{He20, Chen20, Grill2020, Caron20}. This cropping procedure randomly resizes the input \textit{area} between a minimum scale and a maximum scale and jitters aspect ratio between 3/4 to 4/3, before taking a 224$^\text{2}$ crop.
	
	We do not change the cropping for downstream training, as this can drop accuracy significantly (by $\app$2\% on K400). 
	
	In \tblref{tab:crop_aug} we ablate this approach for MoCo (the augmentation in the downstream evaluators are unchanged). 
	
	The first ablation shows the comparison of default cropping \cite{Simonyan15, He16} with a similar version that randomly crops a fraction between $[$0.49, 0.76$]$ $=$  $[$224$^\text{2}$$/$320$^\text{2}$,  224$^\text{2}$$/$256$^\text{2}$$]$ of the original area, instead of the short-side. The performance degrades by 1\% on K400 linear evaluation. Randomly cropping based on area favors larger crops over the short-side resizing and we observe lower training error for this variant. 
	
	Next, adding aspect ratio augmentation can recover some of this performance (65.4\%), and using a smaller minimum area of 0.2, with the maximum area of 0.76 leads to best performance of \textbf{66.8}\%. Using the default values for Inception \cite{Szegedy15} training, $[$0.08,  1.00$]$, appears to be too aggressive.   
	
	\begin{table}[h] 
		\centering
		\tablestyle{1.5pt}{1.05} 
		\begin{tabular}{x{25}|x{26}x{26}|x{26}x{26}}
			& \multicolumn{2}{c|}{\textbf{MoCo} ($\rho{=}$4)} & \multicolumn{2}{c}{\textbf{BYOL} ($\rho{=}$4)}    \\
			\textit{aug+} &   K400 &UCF101 &  K400 &UCF101 \\ 		\shline
			& 67.8 & 93.5 & 68.9 & 93.8 \\ 
			\checkmark	  & 69.0 & 93.6 & 69.8 & 93.9  \\
		\end{tabular}
		\vspace{.3em}
		\caption{\textbf{Stronger augmentations.} Data: K400, 200 epochs. ``\textit{aug+'}' combines the best color and cropping augmentations from \tblref{tab:color_aug} and \tblref{tab:crop_aug}, respectively. }
		\label{tab:enhanced_aug} \vspace{-10pt}
	\end{table}

	\paragraph{Combined augmentations.} We pull together the best color and  cropping augmentations in Tables \ref{tab:color_aug}~\&~\ref{tab:crop_aug}, and train MoCo and BYOL with $\rho{=}$4 for 200ep on K400. The result shown as ``\textit{aug+}'' in \tblref{tab:enhanced_aug} can increase performance on K400 by $\app$1\%. Training the linear classifier of \mbox{{BYOL} ($\rho{=}$4)} for 100ep instead of 60ep leads to our best accuracy of \textbf{70.0}\% on K400, which is 4.7\% below the supervised \mbox{R-50, Slow 8\x8} accuracy of 74.7\%. 
	
	\newcommand{\pacc}[1]{{\bf \fontsize{7.5}{42}\selectfont \color{citecolor!80}~(#1)}}
	\newcommand{\macc}[1]{{\bf \fontsize{7.5}{42}\selectfont \color{lightred!70}~(#1)}}
	\begin{table*}[t] 
		\vspace{-18pt}
		\centering
		\tablestyle{8pt}{1.05} 
		\begin{tabular}{ll|l|llll}
			&&\textit{linear protocol}&\multicolumn{4}{c}{\textit{finetuning accuracy}} \\
			method & pre-train &\textbf{K400} & \textbf{UCF101} & \textbf{AVA} (mAP) & \textbf{Charades} (mAP) & \textbf{SSv2} \\
			\shline
			supervised & \textit{scratch} & \textbf{74.7} & 68.8 & 11.7  & 7.4  & 48.8 \\
			supervised &  K400-240K & - & \textbf{94.8} & 22.2  & \textbf{34.7}  & 52.8 \\ \hline
			\textbf{SimCLR}&  \multirow{4}{*}{K400-240K} & 62.0\macc{$-$12.7}  & 87.9\macc{$-$6.9}  & 17.6\macc{$-$4.6}  & 11.4\macc{$-$23.3}   &  52.0\macc{$-$0.8}       \\ 
			\textbf{SwAV} & & 62.7\macc{$-$11.5}  & 89.4\macc{$-$5.4}  & 18.2\macc{$-$4.0}  & 10.7\macc{$-$24.0}   &  51.7\macc{$-$1.1}       \\ 
			\textbf{BYOL}  && 68.3\macc{$-$6.4}  & 93.8\macc{$-$1.0}  & \textbf{23.4}\pacc{$+$1.2}  & 21.0\macc{$-$13.7}   &  \textbf{55.8}\pacc{$+$3.0}       \\ 
			\textbf{MoCo} & & 67.3\macc{$-$7.4}  & 92.8\macc{$-$2.0}  & 20.3\macc{$-$1.9}  & 33.5\macc{$-$1.2}   &  54.4\pacc{$+$1.8}       \\ \hline
			\textbf{MoCo} & IG-Curated-1M & 69.9\macc{$-$4.8} & \textbf{94.4}\macc{$-$0.4} &  20.4\macc{$-$1.8} &\textbf{34.9}\pacc{$+$0.2} & 54.5\pacc{$+$1.8} \\
			\textbf{MoCo} & IG-Uncurated-1M & 66.0\macc{$-$8.7} & 92.9\macc{$-$2.1} &  20.5\macc{$-$1.7} & 31.3\macc{$-$3.4} & 53.2\pacc{$+$0.4} 
		\end{tabular}
		\caption{\textbf{Downstream benchmarks}: We use linear evaluation on K400 and finetuning accuracy on the other datasets. 200 epochs.  $\rho{=}$3.}
		\label{tab:downstream}
	\end{table*}
	\subsection{Alternative downstream tasks}
	The gap between K400 and UCF101 accuracy  in \sref{sef:backbones} question if solely looking at typical evaluation of UCF101 (or the smaller HMDB51) is enough to identify and rank approaches for unsupervised  learning in video. 
	
	\tblref{tab:downstream} studies several new downstream tasks for unsupervised representation learning in video. We use our MoCo, SimCLR, BYOL and SwAV models trained with  $\rho{=}$3 for 200 epochs on {K400} and evaluate their performance by finetuning on Charades~\cite{Sigurdsson2016}, AVA~\cite{Gu2018},  or Something-Something~\cite{ssv2} (in addition to the K400 linear readout performance and UCF101 performance reported in \tblref{tab:crops_vs_clips}). Details on implementation are given in \S\ref{sec:impl_details}. 
	
	The first two rows in \tblref{tab:downstream} show the two main competitors for this evaluation: (i) training from {scratch} on the datasets and (ii) K400 pre-training. First, we observe that the supervised pre-trained backbones outperform the train-from-scratch counterpart significantly, as expected. 
	
	\paragraph{Downstream datasets.}
	For \textbf{K400} pre-training and linear evaluation, its supervised counterpart has an advantage between 12.7\% and 6.4\% top-1 accuracy among the methods. 
	
	On \textbf{UCF101} unsupervised pre-training is only 1\% lower than the supervised counterpart for BYOL (the strongest). 
	
	On \textbf{AVA} short-term action detection we observe that the BYOL pre-trained model is able to outperform the supervised counterpart by \textbf{+1.2}\% mAP, when using the same, fixed region proposals~\cite{Feichtenhofer19}. This result is significant, as \eg switching from K400 to K600 (nearly double the size of K400) pre-training on AVA leads to a smaller gains in performance~\cite{Feichtenhofer19}.  Overall this is a surprising result as the tasks in K400 and AVA are similar~\cite{Li20}, only that the \textit{temporal} granularity of the actions in AVA is \textit{finer} while their \textit{semantic} granularity is \textit{coarser}; \eg ``\textit{shoot}'' in AVA \vs ``\textit{playing paintball}'' in Kinetics, which might be better captured by the BYOL objective which solely works on positive temporal samples of a video, without contrasting them to other videos (``\textit{shoot}'' might be a positive appearing in many different videos and contrasting them could be harmful to downstream performance). This line of thinking is supported with MoCo's (contrastive objective) performance that is 3.1\% worse than BYOL on AVA. Similarly, SimCLR (contrastive) is worse than SwAV (non-contrastive) when benchmarked on AVA. 
	
	On \textbf{Charades}, long-term action classification, we observe the opposite. Here, the contrastive MoCo is clearly the best performer with 33.5\% mAP (close to the supervised pre-training performance of 34.7\% mAP), while the non-contrastive BYOL is 12.5\% lower. Similarly, now \mbox{SimCLR} (contrastive) is better than SwAV (non-contrastive). Compared to AVA, Charades is a temporally less localized dataset containing activities that need to be recognized from a longer temporal range video, for which contrastive pre-training appears to be outperforming the non-contrastive variants. 
	
	On \textbf{Something-Something v2} (SSv2 in \tblref{tab:downstream}), all the methods perform strong, with BYOL pre-training showing the largest gain of \textbf{+3}\% over supervised pre-training on Kinetics (55.8\% \vs 52.8\% top-1 accuracy).
	
	\paragraph{Pre-training sets: Kinetics \vs IG.}
	Next, we experiment with pre-training on videos from the web. We first investigate \textbf{IG-Curated-1M} \cite{Ghadiyaram2019}, which is a dataset that has been collected with hashtags that are similar to Kinetics labels. This data is a 1M subset of the original 65M introduced in \cite{Ghadiyaram2019}. Using this data (penultimate row in \tblref{tab:downstream}) can excel the performance of MoCo with K400 pre-training, which has a training set of 240K samples (roughly 4.2\x~smaller), and surprisingly even outperforms pre-training on K400 linear readout itself (69.9\% \vs 67.3\% accuracy).
	
	Second, we ablate the effect of using uncurated videos, with \textbf{IG-Uncurated-1M} which are purely random videos taken from the web. On most downstream tasks performance shown in the last row of \tblref{tab:downstream} is equal or only slightly lower than pre-training on K400. Specifically, MoCo changes by \mbox{-1.3}\% on K400 (as expected), +0.1\% on UCF, +0.2\% on AVA, -2.2\% on Charades and -1.2\% on Something-Something v2.  This is an encouraging result for unsupervised learning, as only $\app$4.2\x the number of videos but \textit{random ones}  are required to match the performance of supervised K400 pre-training on the UCF101 and AVA. 
		
	Overall, our results indicate that unsupervised pre-training can be a new paradigm for all of these downstream tasks, for which supervised pre-training is the de-facto standard to achieve best performance. Further, the large difference in performance for pre-training methodologies and objectives (\eg contrastive/non-contrastive) revealed in the light of these benchmarks signals large room for future work.
			
	\begin{table}[h!] 
		\vspace{-8pt}
		\hspace*{-8pt}
		\centering
		\small
		\tablestyle{2.0pt}{1.05}
		\begin{tabular}{l|l|l|r|x{10}|x{15}|x{18}|x{22}| x{22}}
			\multicolumn{1}{c|}{method} &  \multicolumn{1}{c|}{pre-train} &  \multicolumn{1}{c|}{backbone} & param & $T$ & mod & UCF & HMDB & K400 \\   
			\shline
			XDC~\cite{Alwassel19}     & K400  & {\scriptsize R(2+1)D-18} & 15.4M & 32 & V+A & 84.2 & 47.1 \\ 
			GDT~\cite{Patrick20}      & K400   &  {\scriptsize R(2+1)D-18} & 15.4M& 32 &V+A & 89.3 & 60.0  \\
			MMV \cite{alayrac2020self}    & AS+HT & S3D-G & 9.1M & 32 & {\tiny V+A+T} & 92.5 & 69.6 \\ \cline{1-8}

			SpeedNet~\cite{Benaim20}   &  K400 & S3D-G & 9.1M & 64    & V  &  81.1 & 48.8  \\
			CoCLR   \cite{Han20}        & K400    & S3D-G & 9.1M & 32    & V & {87.9} &  {54.6}   \\
			CoCLR   \cite{Han20}      & K400    & 2\x S3D-G  & 9.1M & 32   & V & {90.6} &  {62.9}   \\
			VTHCL~\cite{yang2020video}     & K400 & R-50  & 31.8M  &  8 & V & 82.1 & 49.2 & 37.8 \\  	
			CVRL~\cite{Qian20}    & K400 & R-50  & 31.8M  &  32 & V & 92.2 & 66.7 & 66.1 \\  	
		
			\shline
			\textbf{$\rho$BYOL}  & K400 & R-50  & 31.8M &  8 & V & \textbf{94.2} & \textbf{72.1} & \textbf{70.0} \\
			\textbf{$\rho$BYOL}  & K400 & R-50  & 31.8M &  16 & V & \textbf{95.5} & \textbf{73.6} & \textbf{71.5}  \\ 
			\textbf{$\rho$BYOL}  & K400 &  {\scriptsize R(2+1)D-18} & 15.4M &  32 & V & \textbf{94.4} & \textbf{72.2}  \\
			\textbf{$\rho$BYOL}  & K400 & S3D-G & 9.1M &  32 & V & \textbf{96.3} & \textbf{75.0}  
		\end{tabular}
		\caption{\textbf{Comparison with state-of-the-art}. ``param'' indicates the number of parameters, $T$ inference frames, in the backbone. ``V'' is Vision, ``A'' is Audio, ``T'' Text modality. \textbf{$\rho$BYOL} is our best model trained with temporal persistency of $\rho{=}$4. We report fine-tuning accuracy on UCF/HMDB and linear accuracy on K400.}
		\label{tab:sota:ucf_hmdb} \vspace{-10pt}
	\end{table} 	

	\subsection{Comparison to previous work}
	
	In a final experiment we take the best model from \tblref{tab:enhanced_aug} and compare it with the state-of-the-art using the commonly used protocols on UCF101 and HMDB51 (across all 3 train/val splits) and K400. In \tblref{tab:sota:ucf_hmdb} we show the results. 
	
	The strongest previous approaches are using multi-modal input, Vision ``V'', Audio ``A'', Text ``T'', to train a contrastive objective across modalities;  XDC~\cite{Alwassel19} performs DeepCluster \cite{Caron18} on (V+A), CVRL~\cite{Qian20}, GDT~\cite{Patrick20} and MMV~\cite{alayrac2020self}  use an objective similar to \mbox{SimCLR} on (V), (V+A), and (V+A+T), with the latter training on a Audioset (AS) \cite{gemmeke2017audio} and HowTo100M (HT) \cite{miech2019howto100m}, and CoCLR~\cite{Han20} can be seen as a variant of MoCo on rgb and optical-flow input. 
	
	In comparisons, our best performing model \textbf{$\rho$BYOL}, which is BYOL trained with temporal persistency over $\rho{=}$4 clips,  (\textit{cf.}~Tables~\ref{tab:crops_vs_clips}~\&~\ref{tab:enhanced_aug}), provides a substantial performance gain over the best published method \cite{Han20}: \textbf{+5.7}\% and \textbf{+12.1}\% top-1 accuracy on UCF101 and HMDB51 (using identical backbone and pre-training data). 
	
	On K400 linear evaluation with the same data and  R-50, Slow pathway \cite{Feichtenhofer19} as backbone, our approach outperforms the concurrent CVRL~\cite{Qian20} by \textbf{+5.4}\% accuracy.
		
	\section{Conclusion}

	This paper has studied four meta-methodologies for unsupervised learning from video. Our findings include that it is beneficial to sample positives with longer timespans between them, contrastive objectives are less influential than momentum encoders, and training duration, backbones, video augmentation and curation are all critical for good performance. 
	Our resulting models which learn persistent features across augmented spacetime clips set a new state-of-the-art. 
	
	We observed that linear readout on Kinetics  is a good indicator of the performance on other datasets and that unsupervised pre-training can compete with the supervised counterpart on several datasets, but there is  room for improvement. 		
	We hope that our baselines will foster research and provide common ground for future comparisons. 

	\appendix
	\cleardoublepage

	This appendix provides additional material:
	\S\ref{sec:extra_results} contains further results on ``in-the-wild'' data (\S\ref{sec:extra_IG}), Kinetics-600 (K600)~\cite{carreira2018short} and Kinetics-700 (K700)~\cite{Carreira19} data (\S\ref{sec:extra_Kinetics}) and on the effect of key implementation details (\S\ref{sec:results_impl}).
	
	\S\ref{sec:impl_details} contains additional implementation details  for: Unsupervised pre-training (\S\ref{sec:training}), and downstream evaluation in Kinetics (\S\ref{sec:kinetics}), AVA   (\S\ref{sec:detection}),  Charades (\S\ref{sec:charades}), Something-Something V2 (\S\ref{sec:ssv2}), UCF101 (\S\ref{sec:ucf}), HMDB51 (\S\ref{sec:hmdb}).

	\section{Additional Results} \label{sec:extra_results}
	\subsection{Scaling ``in-the-wild'' data} \label{sec:extra_IG}
	
	As a follow-up experiment \tblref{tab:ep-cs-delta-byol} compares training BYOL longer (200ep) to increasing its clips-size $\rho$ but not training longer (50ep). For both \subref{tab:ep-cs-delta-byol-IG-kin} curated and \subref{tab:ep-cs-delta-byol-IG-R} random data, this results in a significant gain of performance. 
	
	\begin{table}[h] \vspace{-10pt}
		\centering
		\captionsetup[subfloat]{captionskip=2pt}
		\captionsetup[subffloat]{justification=centering}
		\subfloat[\textbf{IG-Curated-1M}.
		\label{tab:ep-cs-delta-byol-IG-kin}]{
			\tablestyle{1pt}{1.05} 
			\hspace{-15pt}
			\begin{tabular}{l|r|x{26}x{26}}
				\multicolumn{2}{c}{} 	&  \multicolumn{2}{c}{ \textbf{BYOL} }    \\
				\multicolumn{1}{c|}{$\rho$} & ep  & K400 &UCF101  \\
				\shline
				2 & 50 & 64.1	& 93.5  \\ 
				2 & 200  & 60.2 & 92.7   \\
				4 & 50 &  67.7 &	94.5 \\ 
		\end{tabular}} \qquad
		\subfloat[\textbf{IG-Uncurated-1M}. 	\label{tab:ep-cs-delta-byol-IG-R} ]{
			\tablestyle{1pt}{1.05} 
			\begin{tabular}{l|r|x{26}x{26}}
				\multicolumn{2}{c}{} 	&  \multicolumn{2}{c}{ \textbf{BYOL} }    \\
				\multicolumn{1}{c|}{$\rho$} & ep  & K400 &UCF101  \\
				\shline
				2 & 50 &   58.9	& 90.1 \\ 
				2 & 200 &  57.9 & 91.6    \\
				4 & 50  & 63.8 &	91.8  \\ 
		\end{tabular}}   \hspace{-15pt}
		\caption{\textbf{More epochs (ep) \vs more clips ($\rho$)},  Longer training degrades performance for BYOL, but increasing $\rho$ does not. } 
		\label{tab:ep-cs-delta-byol} 
	\end{table}
	
	We also explore an experiment for increasing the clip-size in MoCo and training longer (as MoCo works stable for more epochs). \tblref{tab:ep-vs-delta-IG-kin} shows the results. It can be observed that increasing the number of clips from $\rho{=}$2 to $\rho{=}$3 can increase the results by 1.6\%/0.9\% K400 and 0.4\%/1\% on UCF101 for 100/200ep training. Going to $\rho=4$ brings further gain.  In terms of efficiency, increasing $\rho$ is both more accurate and faster than increasing the number of epochs, \eg training MoCo ($\rho{=}$3, 100ep) takes only 63\% of the duration that {MoCo} ($\rho{=}$2, 200ep) requires.
	
	\begin{table}[h] 
		\centering
		\tablestyle{1pt}{1.05} 
		\begin{tabular}{x{14}|x{26}x{26}|x{26}x{26}|x{26}x{26}}
			&  \multicolumn{2}{c|}{ \textbf{MoCo} ($\rho{=}$2)}  &   \multicolumn{2}{c|}{ \textbf{MoCo} ($\rho{=}$3)} &  \multicolumn{2}{c}{ \textbf{MoCo} ($\rho{=}$4)}  \\
			\multicolumn{1}{c|}{ep}  & K400 &UCF101 & K400 &UCF101 & K400 &UCF101  \\
			\shline
			100 &  67.5	& 93.3 & 69.1 &	93.7 & 69.8 & 94.9  \\ 
			200  &  69.0 & 93.4 & 69.9 &	94.4 & 69.9 & 94.9 \\
		\end{tabular}
		\vspace{.3em}
		\caption{\textbf{More epochs (ep) \vs more clips ($\rho$)}:  Dataset: \textbf{IG-Curated-1M},  $\rho{=}$2. Training longer is less effective than increasing the number of temporal clips per iteration ($\rho$).}
		\label{tab:ep-vs-delta-IG-kin}
	\end{table}
		
	Finally, we remark that the IG-Curated-1M is subsampled such that the hastags are uniformly distributed (roughly balanced). Therefore this dataset is matching K400 in terms of content and distribution. We revisit this point next by investigating the effect of scale, curation and balancing of the video data. 
	
\begin{figure}[t!]
	\centering
	\vspace{-0.5em}
	\includegraphics[width=1.0\linewidth]{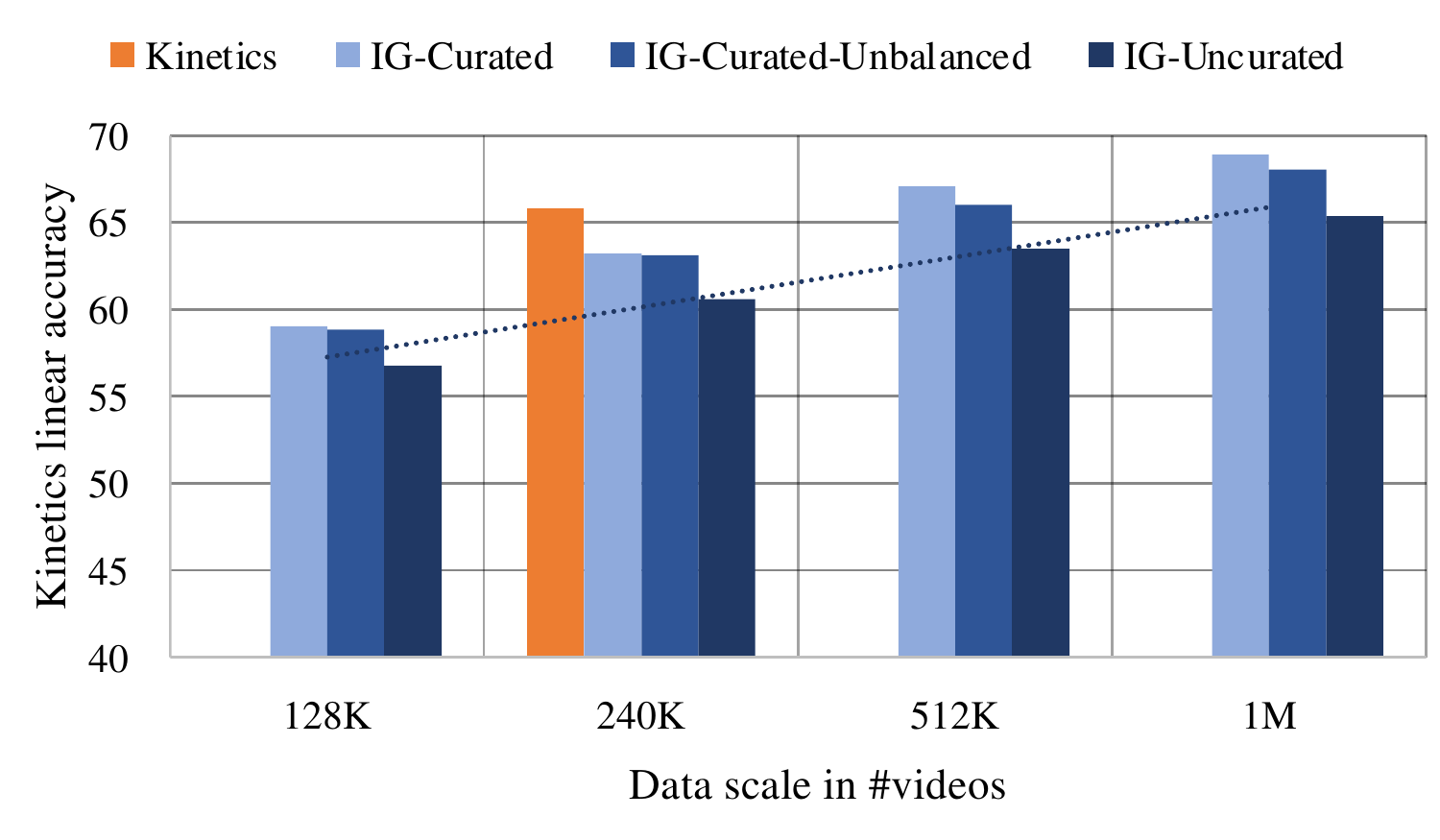}
	\caption{\textbf{Data scale and curation.} We increase dataset size (number of videos) for IG-Curated,  IG-Curated-Unbalanced, and IG-Uncurated.  By using 4\x~the number of videos, IG-Uncurated approaches the heavily curated Kinetics (K400) pre-training on K400 linear evaluation protocol. The dotted line represents a linear trend. Method: \textbf{MoCo}, 200 epochs, $\rho{=}$2. 
		\vspace{-1em}
	}
	\label{fig:data_scale}
\end{figure}

	In this experiment, we increase the scale of the data from 128K to 1M distinct videos.
	We increase dataset size (number of videos) for IG-Curated \cite{Ghadiyaram2019},  IG-Curated-Unbalanced \cite{Ghadiyaram2019} (which has random class distribution), and IG-Uncurated (which are random IG videos).   The experiment with 200-epoch {MoCo} with $\rho{=}$2, linear protocol downstream evaluation on K400 is shown in \figref{fig:data_scale} and reveals: 
	
	(i) Comparing the curation axis: At 240K training samples, the four data sources provide 65.8\%, 63.2\%,	63.1\%, 60.6\% top-1 accuracy for K400, IG-Curated, IG-Curated-Unbalanced and IG-Uncurated, respectively. The decay from the heavily curated K400 to IG-Curated (2.6\%) is similar to the one from IG-Curated to IG-Uncurated (2.5\%), while the class balancing seems to have a minor effect on accuracy. 
	
	(ii) Comparing the scale axis: Doubling the data scale (number of videos) roughly linearly increases the accuracy across all datasets. With 1M uncurated videos the performance approaches 65.4\% which is similar to the 65.8\% produced by using K400 pre-training. The experiment indicates that it is possible to approach unsupervised Kinetics pre-training when using 4\x more (1M \vs 240K in Kinetics), but \textit{random}, videos when evaluating on Kinetics. 

	\subsection{Scaling Kinetics data}  \label{sec:extra_Kinetics}

\begin{table*}[!h] 
	\centering
	\vspace{-15pt}
	\tablestyle{8pt}{1.05} 
	\begin{tabular}{x{50}|x{26}x{26}|x{26}x{26}x{26}|x{26}}
		& \multicolumn{2}{c|}{pre-train} & \multicolumn{3}{c|}{} & \textit{finetune} \\  
		method &  data & \#videos &\textbf{K400} & \textbf{K600} & \textbf{K700} & \textbf{UCF101} \\
		\shline
		\multirow{2}{*}{\textit{supervised}}  & \multicolumn{2}{c|}{\textit{scratch}}  & \textbf{74.7} & \textbf{78.1} & \textbf{65.2} & 68.8 \\ \cline{2-7}
		&  K400 & 240k  & \multicolumn{3}{c|}{\textit{linear protocol}} & \textbf{94.8}  \\ \shline
		\textbf{MoCo} ($\rho{=}$4) & K400 & 240k & 69.0 &  70.0 &  54.2    & 93.6    \\  \hline
		\textbf{MoCo} ($\rho{=}$2) & \multirow{2}{*}{K600} &  \multirow{2}{*}{387k}   & 69.6  &  70.7 & 55.1 & 92.7     \\ 
		\textbf{MoCo} ($\rho{=}$4)  &  & &71.5 &  72.8  & 57.7  & 94.5   \\ \hline
		\textbf{MoCo} ($\rho{=}$2)  & \multirow{2}{*}{K700}  &  \multirow{2}{*}{522k} & 70.0 &  71.4  & 56.2   &  92.8     \\  
		\textbf{MoCo} ($\rho{=}$4) &  &  & 71.7 & 73.2 & 58.1   & \textbf{94.8}    \\

	\end{tabular}
	\vspace{0.5em}
	\caption{\textbf{Dataset scale}:  Configuration: backbone: R-50, Slow 8 \x~8, 200 epochs. Our approach, \textbf{MoCo} ($\rho{=}$4), is able to approach supervised pre-training on the popular UCF101 evaluation protocol, but there remains a gap for the linear protocol on K400, K600 and K700.  }
	\label{tab:k600_k700}
	\vspace{-10pt}
\end{table*}

As referenced in Sec.~{\color{red}4} of the main paper, \tblref{tab:k600_k700} shows a series of extra results for pre-training on the larger-scale Kinetics-600 (K600)~\cite{carreira2018short} and Kinetics-700 (K700)~\cite{Carreira19} datasets, and is analyzed next: The first row of the table shows supervised training on the respective datasets, where UCF101 has two entries, one for training-from-scratch and one for using K400 as pre-training. 

For the experiments we focus on our temporally persistent MoCo algorithm and, as in the main paper, evaluate Kinetics with the linear classification protocol and UCF101 by finetuning all weights.  
The first unsupervised row in \tblref{tab:k600_k700} shows our best \textbf{K400} pre-trained \textbf{MoCo} ($\rho{=}$4) model, achieving 69.0\%, 70.0\%, 54.2\% and 93.6\% on K400, K600, K700 and UCF101, respectively (this is the model with strong augmentations from Table {\color{red}10} of the main paper). 

The next row shows MoCo trained on \textbf{K600} with a temporal persistency objective across two clips, $\rho{=}$2. This version is able to slightly outperform the K400 pre-trained variant on all datasets, except UCF101. Directly comparing this version with learning temporal persistency across $\rho{=}$4  clips can significantly increase accuracy on all datasets by $\app$2\%. 

The final two rows of \tblref{tab:k600_k700}, show the same two models when pre-trained on \textbf{K700}. Here, we see that going from K400 to K700 increases accuracy by  2.7\%, 3.2\% and 3.9\%, 1.2\% on K400, K600, K700 and UCF101, respectively.

Overall the experiments suggest clear \textit{benefits of using larger-scale datasets} for unsupervised pre-training and room for improvement under the linear classification protocol, especially when evaluated on larger datasets.

\subsection{Key implementation specifics} \label{sec:results_impl}
While the full implementation details of all four meta-methodologies are provided in \S\ref{sec:training}, we want to discuss the most impactful ones, which we found critical to achieve good performance in their realizations, throughout this section.

\begin{table}[h!] 
	\centering
	\tablestyle{3pt}{1.05} 
	\tablestyle{6.4pt}{1.0}	
	\begin{tabular}{ccccccccc}
		$m_\text{base}$ & N/A & 0.988 & 0.990 & 0.992  & 0.994 & 0.996 \\
		\shline
		acc. & 64.5 & 65.5  & 65.5 & 65.6 & 65.8 & 65.1 \\
	\end{tabular}
	\vspace{.3em}
	\caption{\textbf{Momentum annealing for MoCo}. Dataset: \textbf{K400}, 200 epochs, $\rho{=}$ 2. Using cosine-annealing of the momentum brings gains of $\app$1\% accuracy. We use 0.994 as default for MoCo. }
	\label{tab:mom_moco} \vspace{-10pt}
\end{table}

\paragraph{Momentum annealing.} BYOL is using an annealing of the rate at which parameters of the momentum encoder $\theta_m$, that are a moving average, with momentum $m$, of the trained encoder $\theta$. During training BYOL starts with a momentum of $m_\text{base}{=}$0.996 and increases it to 1 with a cosine annealing $m = 1 - (1-m_\text{base})\cdot ({\cos ({\pi k / K}) + 1}) / 2$ with $k$ the current iteration and $K$ the maximum number of training iterations~\cite{Grill2020} (this is unrelated to the learning rate decay).

By default MoCo, is using a fixed momentum of $m=0.999$ during training. In \tblref{tab:mom_moco}, we ablate the positive (or negative) effect of using momentum annealing with different starting rates $m_\text{base}$ for MoCo. We observe that not using any annealing (N/A) produces 64.5\% accuracy and using momentum annealing can boost this performance by $\app$1\%, while being relatively stable for different values of $m_\text{base}$. Consequently, we are using momentum annealing with \mbox{$m_\text{base}=0.994$} for all our MoCo experiments. 

\begin{figure}[h!]
	\centering
	\includegraphics[width=0.9\linewidth]{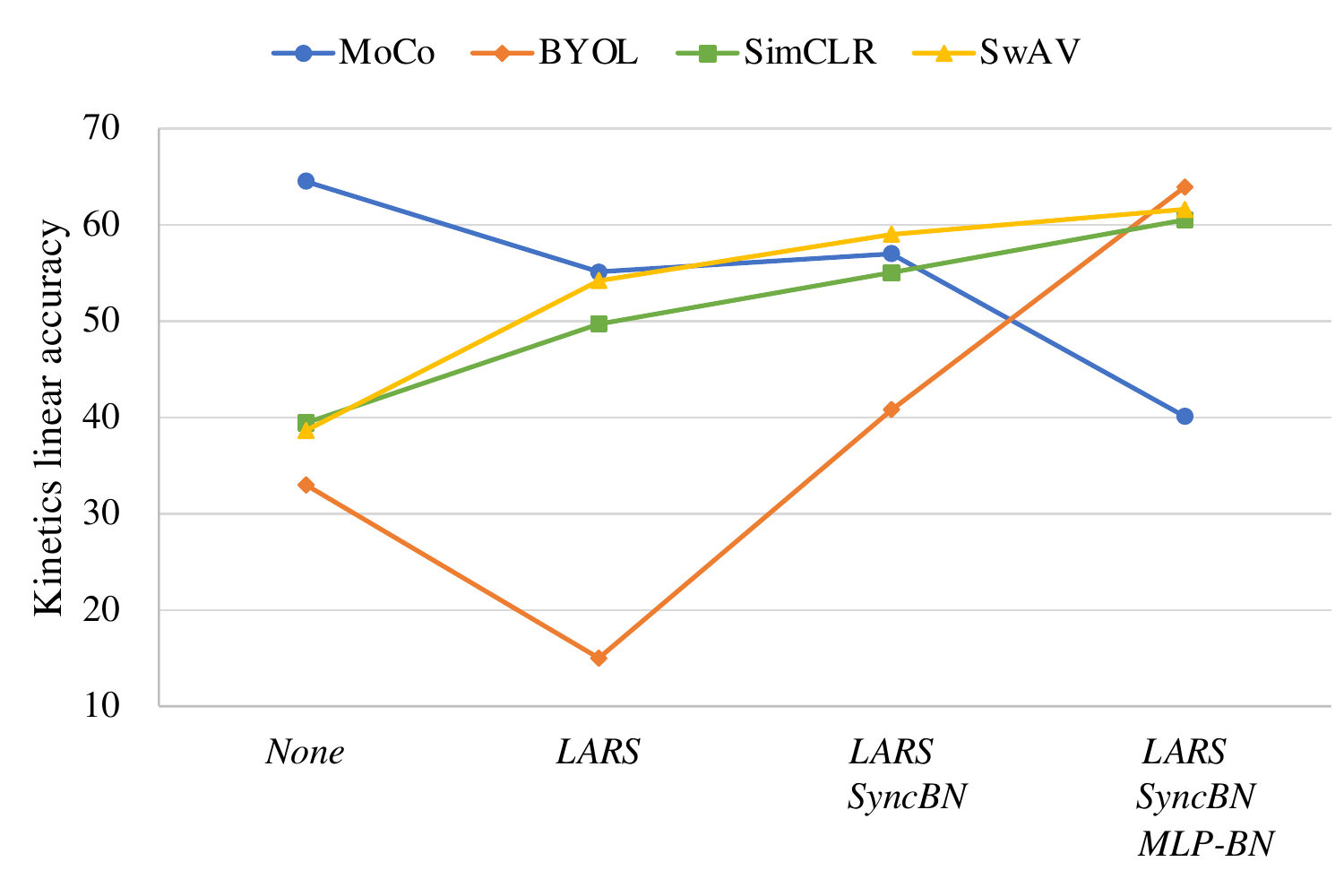}
		\vspace{-5pt}
	\caption{\textbf{Key implementation specifics.}  BYOL, SimCLR, SwAV heavily rely on \textit{LARS}, \textit{SyncBN}, and BN in the MLP (\textit{MLP-B}N), MoCo does not require these, but  does not benefit of having them. 
	}
	\label{fig:syncBN}
		\vspace{-10pt}
\end{figure}

\paragraph{Normalization and optimization.}

Here, we present normalization specifics that we found critical to achieve good performance in the underlying implementation of the methods: SimCLR, BYOL and SwAV are using synchronized Batch-Normalization (BN) \cite{Ioffe15} statistics (\textit{SyncBN}) across 8 GPUs during training, batch-normalization after every MLP layer (\textit{MLP-BN}), and a large-batch optimizer (\textit{LARS}) \cite{you2017large}. LARS adaptively scales the learning rate for each individual parameter by using the ratio between gradient and parameter magnitudes.
MoCo is not using these components (\textit{None}) by default. 
In \figref{fig:syncBN} we illustrate the results. It shows accuracy on K400 linear readout, if step-by-step adding these specifics to the methods. We make the following observations:

(i) Using None of the augmentations provides best performance for MoCo (its default) but significantly degrades BYOL, SimCLR and SwAV. Here, it is worth noting that BYOL provides decent accuracy of 32.9\% without \textit{SyncBN}, \textit{LARS} and any BN in the MLP.

(ii) Adding \textit{LARS} optimizer reduces performance in MoCo and BYOL, while having a boost of around 10\% for both SimCLR and SwAV. It is interesting, that solely using a more advanced optimizer, which adapts the learning rates of the weights according to their gradient magnitudes, decreases performance in methods using a momentum encoder (MoCo, BYOL), but boosts it without (SimCLR, SwAV).

(iii) further adding \textit{SyncBN} and \textit{MLP-BN} increases BYOL performance dramatically; this related to recent studies \cite{richemond2020byol} which suggest that {normalization} is important to achieve good performance using BYOL. 

(iv) While BYOL, SimCLR and SwAV do show further gains for adding \textit{SyncBN} and \textit{MLP-BN}, MoCo shows no significant change for using \textit{SyncBN}, and degrades drastically in performance for using BN in the MLP-head. 

\paragraph{Projection MLP.} It has been shown that using a deeper projection MLP in pre-training can increase the accuracy of the resulting representations for image classification~\cite{Chen20,Chen20moco,chen2020big}. Here, we investigate the effect of more hidden layers for video classification, across all four meta architectures. The results are shown in \tblref{tab:mlp} and discussed next.   

(i) MoCo achieves a significant gain of 1.2\% on K400 for using a 3-layer (2 hidden layers) MLP \vs using a 2-layer MLP and there is no gain for using a 4$^\text{th}$ layer. UCF performance appears stable to this modification. The gain is in line with results in image classification~\cite{Chen20moco}.

(ii) For BYOL, which has an additional \textit{Predictor MLP}, with weights ${\theta_p}$ (see \figref{fig:methods:byol}), we ablate two dimensions: increasing the projection depth, and the prediction depth. Our results show that using 3-layer projection \vs 2-layer does not affect performance on K400, and has a decay of -0.7\% on UCF101. Increasing also the depth of the predictor from our default value of 2 to 3 layers will lead to a significant decrease of -2.2\% and -2.5\% on both K400 and UCF101. 

(iii) SimCLR, shows similar behavior as MoCo: A consistent gain for using 3 projection layers (+1.5\% on K400, +0.5\% on UCF101), and no further gain for a 4-layer MLP.

(iv) SwAV shows continuing gains on K400 for adding more MLP layers, +1.3\% for going from 2 to 3 and another +0.4\% for 4-layer MLP; however, its UCF-101 performance is decaying with more projection layers. 

Overall,  \tblref{tab:mlp}  suggests that K400 linear evaluation accuracy gernally benefits from deeper projection heads, while the performance for fine-tuned UCF101 downstream performance is relatively unchanged and rather shows a decaying effect for deeper MLPs. 
When studying the training complexity for pre-training, which we measure as floating point operations (FLOPs) and Parameters for the full training architecture (encoders + MLPs),  \tblref{tab:mlp}  shows that FLOPs are mostly unchanged by deeper MLPs (as they operate on feature maps of size 1\x 1\x 1), but parameters increase leading to large models especially for momentum encoder based approaches (MoCo and BYOL). 

\begin{table}[t!] 
	\centering
	\tablestyle{3pt}{1.05} 
	\tablestyle{6.4pt}{1.0}	
	\vspace{-10pt}
	\begin{tabular}{l|c|cc|cc}
		\multicolumn{1}{c|}{\multirow{2}{*}{method}} & \multirow{2}{*}{MLP layers} & \multicolumn{2}{c}{training} & \multicolumn{2}{c}{accuracy} \\
		& & FLOPs & Param & K400 & UCF101  \\
		\hline
		\multirow{3}{*}{MoCo} & 2 & 41.74G & 72.2M & 64.6 & \textbf{91.3} \\
		& 3 & 41.74G & 80.6M & \textbf{{65.8}} &\textbf{{91.0}} \\
		& 4 & 41.75G & 88.9M  & 65.7 & 91.0 \\
		\hline
		\multirow{3}{*}{BYOL} & 2, predictor: 2 & 41.75G & 86.4M & {\textbf{65.8}} & \textbf{{92.7}} \\
		& 3, predictor: 2 & 41.77G &  119.9M & {65.8} & 92.0  \\
		& 3, predictor: 3 & 41.78G &  153.5M & 63.6 & 90.2  \\
		\hline
		\multirow{3}{*}{SimCLR} & 2 & 41.74G & 36.1M & 59.0	& 88.4 \\
		& 3 & 41.75G & 40.3M & 60.5	& \textbf{{88.9}} \\
		& 4 & 41.75G & 44.5M  & \textbf{{60.6}}	& 88.5 \\
		\hline
		\multirow{3}{*}{SwAV} & 2 & 41.74G & 36.2M & 60.3 &	\textbf{88.1} \\
		& 3 & 41.75G & 40.4M & 61.6 & 87.3 \\
		& 4 & 41.75G & 44.6M  & \textbf{62.0} &87.1 \\
	\end{tabular}

	\caption{\textbf{Varying depth of MLPs.} Dataset: \textbf{K400}, 200 epochs, $\rho{=}$2. Training complexity is measured in floating point operations (FLOPs) and Parameters. Accuracy is reported as linear evaluation (K400) and fine-tuning (UCF101) of the backbone without MLPs. }
	\label{tab:mlp}
		\vspace{-.5em}
\end{table}

\section{Additional Implementation Details} \label{sec:impl_details} 

\subsection{Unsupervised pre-training}   \label{sec:training} 

\paragraph{Training details.}
We use the initialization outlined in \cite{he2015delving}. The projection and prediction MLP weights are initialized with~\cite{Glorot11}.  We optimize with synchronized SGD training on 64 GPUs with a mini-batch size of 8 clips per GPU; therefore, the total mini-batch size is 512.	
We train with Batch Normalization (BN) \cite{Ioffe15}, and the BN statistics are computed within each 8 clips for MoCo and 64 clips by synchronizing across 8 GPUs (\textit{SyncBN}) for BYOL, SimCLR and SwAV.
We adopt a half-period cosine schedule \cite{Loshchilov2016} of learning rate decaying: the learning rate at the $n$-th iteration is $\eta\cdot0.5[\cos(\frac{n}{n_\text{max}}\pi)+1]$, where $n_\text{max}$ is the maximum training iterations and the base learning rate $\eta$ is set for each method to $\eta_\text{MoCo}=0.4$, and $\eta_\text{SimCLR} = \eta_\text{BYOL} = \eta_\text{SwAV}=4.8$. We apply (\textit{LARS}) \cite{you2017large} (except for bias and BN parameters \cite{Grill2020}), with trust coefficient of 0.001, for BYOL, SimCLR, and SwAV training. The SGD weight decay is $10^{-4}$ for MoCo and $10^{-6}$ for for BYOL, SimCLR  and SwAV.  The temperature parameter $\alpha=0.1$ for MoCo, SimCLR and SwAV. The projection MLP output dimensions are \mbox{$d_\text{MoCo} = d_\text{SimCLR}  = \eta_\text{SwAV}=128$}, and $d_\text{BYOL}=256$, as in their original publications~\cite{He20,Chen20,Caron20,Grill2020}.

\paragraph{MoCo details.} We use a queue storing 65536 negatives and shuffling BN to avoid intra-batch communication among samples~\cite{He20}. We use a 3-layer (2 hidden layers, ablation in  Table~{\color{red}6} of the main paper) projection MLP with hidden dimension 2048, ReLU activation~\cite{Nair10} and no BN. Other hyperparameters are as in \cite{He20,Chen20moco}. The momentum encoder weights  $\theta_m$ are updated with an annealed momentum  \mbox{$m = 1 - (1-m_\text{base})\cdot ({\cos ({\pi k / K}) + 1}) / 2$} with $k$ the current iteration and $K$ the maximum number of training iterations~\cite{Grill2020}, starting with $m_\text{base} = 0.994$. The corresponding ablation is in  Table~{\color{red}3} of the main paper.

\paragraph{BYOL details.} Our BYOL implementation uses a momentum annealing starting from $m_\text{base} = 0.996$. We minimize the negative cosine similarity in equation ({\color{red}2}) of the main paper multiplied by 2 which is equivalent to BYOL's MSE of $\ell_2$-normalized vectors~\cite{Grill2020}. The projection and prediction MLPs have 2 layers (one hidden layer with dimension 4096) and use BN following the original publication~\cite{Grill2020}.

\paragraph{SimCLR details.} We follow the default implementation~\cite{Chen20}. We use a 3-layer projection MLP with  a hidden dimension of 2048, ReLU and BN. The loss in equation ({\color{red}1}) of the main paper is computed synchronized over the full batch size.

\paragraph{SwAV details.} We follow the default implementation~\cite{Caron20}, using 3 Sinkhorn-Knopp iterations~\cite{cuturi2013sinkhorn} and freezing the prototypes for the first epoch. 
The Sinkhorn regularization parameter is set to $0.05$. As in the default implementation~\cite{Caron20}, the matrix normalization statistics of the Sinkhorn-Knopp algorithm are computed synchronized over the full training batch. 
The projection MLP uses ReLU and BN and is identical to the one used in~\cite{Caron20}, only that we use a 3-layer MLP instead of 2 (ablations are in Table~{\color{red}6} of the main paper).

\paragraph{Encoder details.} Our default encoder,  $f_\theta$, is a R-50 Slow model~\cite{Feichtenhofer19}, \ie a ResNet-50~\cite{He16} with a temporal dimension of size $T$ and sample rate $\tau$. We perform all ablations with default $T$\x $\tau$ of 8\x 8. We show the architecture in~\tblref{tab:arch}. 

\paragraph{Augmentation details.}
We perform video decoding and data augmentation using PyTorch's torchvision package. 

We obtain different clips from a video by the following procedure.
For the temporal dimension, we randomly sample a clip (of $T$\x$\tau$ frames) from the full-length video, and the input to the ResNet encoder are $T$ frames subsampled from the raw clip with a stride of $\tau$; for the spatial dimension, we randomly crop 224\x224 pixels from a video, or its horizontal flip, with a shorter side randomly sampled in [256, 320] pixels \cite{Feichtenhofer19} (VGG-style~\cite{Simonyan15,He16} spatial cropping, a comparison to Inception-style~\cite{Szegedy15} cropping, which we use for results in \S\ref{sec:results_augmentations},  is given in Table~{\color{red}9} of the main paper). 

To each clip, we apply a random horizontal flip, color distortion and Gaussian blur following the SimCLR and MoCo v2 implementation~\cite{Chen20,Chen20moco}.
For color augmentation we use the \texttt{ColorJitter} (probability $0.8$) and \texttt{RandomGrayscale} (probability $0.2$) method from \texttt{torchvision.transforms} module of PyTorch with the color strength parameter $s$: \{brightness, contrast, saturation, hue\} $=$ \{0.4$s$, 0.4$s$, 0.4$s$, 0.1$s$\} By default $s{=}0.5$. Ablations are given in Table~{\color{red}8} of the main paper. For Gaussian blur we use a spatial kernel with standard-deviation $\in [0.1, 2.0]$ applied with probability of $0.5$.

\newcommand{\blocks}[3]{\multirow{3}{*}{\(\left[\begin{array}{c}\text{1$\times$1$^\text{2}$, #2}\\[-.1em] \text{1$\times$3$^\text{2}$, #2}\\[-.1em] \text{1$\times$1$^\text{2}$, #1}\end{array}\right]\)$\times$#3}
}
\newcommand{\blockt}[3]{\multirow{3}{*}{\(\left[\begin{array}{c}\text{\underline{3$\times$1$^\text{2}$}, #2}\\[-.1em] \text{1$\times$3$^\text{2}$, #2}\\[-.1em] \text{1$\times$1$^\text{2}$, #1}\end{array}\right]\)$\times$#3}
}
\newcommand{\outsizes}[7]{\multirow{#7}{*}{\(\begin{array}{c} \text{\emph{Slow}}: \text{#1$\times$#2$^\text{2}$}\\[-.1em] \text{\emph{Fast}}: \text{#4$\times$#5$^\text{2}$}\end{array}\)}
}
\newcommand{\outsizesRaw}[4]{\multirow{#4}{*}{\(\begin{array}{c}  \text{#1$\times$#2$^2$}\\[-.1em]  \end{array}\)}}

\begin{table}[t!]
	\vspace{-10pt}
	\scriptsize
	\centering
	\tablestyle{1pt}{1.08}
	\begin{tabular}{c|c|c}
		stage &  kernels  &  output sizes $T$\x$S^2$ \\
		\shline
		\multirow{1}{*}{raw clip} & - & \outsizesRaw{$T\tau$}{224}{3}{1}  \\
		\hline
		\multirow{2}{*}{data layer} & \multirow{2}{*}{stride $\tau$, 1$^\text{2}$} &   \outsizesRaw{$T$}{224}{3}{2}  \\
		&  &  \\
		\hline
		\multirow{2}{*}{conv$_1$} & \multicolumn{1}{c|}{1\x7$^\text{2}$, {64}} &   \outsizesRaw{$T$}{112}{64}{2}   \\
		& stride 1, 2$^\text{2}$   \\
		\hline
		\multirow{2}{*}{pool$_1$}  & \multicolumn{1}{c|}{1\x3$^\text{2}$ max} &   \outsizesRaw{$T$}{56}{64}{2}  \\
		& stride 1, 2$^\text{2}$  \\
		\hline
		\multirow{3}{*}{res$_2$} & \blocks{{256}}{{64}}{3}  &  \outsizesRaw{$T$}{56}{256}{3}   \\
		&  & \\
		&  & \\
		\hline
		\multirow{3}{*}{res$_3$} & \blocks{{512}}{{128}}{4}  &  \outsizesRaw{$T$}{28}{512}{3}  \\
		&  & \\
		&  & \\
		\hline
		\multirow{3}{*}{res$_4$} & \blockt{{1024}}{{256}}{6}  & \outsizesRaw{$T$}{14}{1024}{3} \\
		&  & \\
		&  & \\
		\hline
		\multirow{3}{*}{res$_5$} & \blockt{{2048}}{{512}}{3}  &   \outsizesRaw{$T$}{7}{2048}{3} \\
		&  & \\
		&  & \\
		\hline
		pool$_5$ & \multicolumn{1}{c|}{global average pool} & \outsizesRaw{1}{1}{2048}{1}  \\
	\end{tabular}
	\vspace{.1em}
	\caption{\textbf{R-50, Slow pathway}~\cite{Feichtenhofer19}. 
		The dimensions of kernels are denoted by $\{$$T$\x $S^2$, $C$$\}$ for temporal, spatial, and channel sizes.
		Strides are denoted as $\{$temporal stride, spatial stride$^2$$\}$.
		Non-degenerate temporal filters are underlined.
		Residual blocks are in brackets. Temporal pooling is only performed at the last layer, collapsing spacetime dimensions. By default $T$\x $\tau$ = 8\x 8.
	}
	\label{tab:arch}
	\vspace{-.8em}
\end{table}

\subsection{Details: Kinetics Action Classification}\label{sec:kinetics}

\paragraph{Datasets.} Kinetics-400 \cite{Kay17} consists of $\app$240k training videos and 20k validation videos in 400 human action categories.
Kinetics-600 \cite{carreira2018short} has $\app$392k training videos and 30k validation videos in 600 classes. 
Kinetics-700 \cite{Carreira19} has $\app$523k training videos and 35k validation videos in 600 classes.

\paragraph{Linear classification protocol.}
We validate the methods by linear classification on frozen features, following the common protocol in image classification~\cite{He20}.
After unsupervised pre-training on Kinetics, we freeze the features of the encoder and train a linear classifier on top of the last layer features (\eg pool$_5$ in \tblref{tab:arch}). For all ablations in the paper the classifier is trained for 60 epochs (using 100 epochs will increase accuracy by $\app$0.2\%) using the same cosine schedule as for pre-training (\sref{sec:training}) with a base learning rate of  $\eta=4.0$ (10\x higher than in pre-training), linear warm-up in the first 8 epochs, and weight decay of 0.  

\paragraph{Training augmentation.} 	
We use the default training augmentation~\cite{Feichtenhofer19}. We randomly sample a clip (of $T$\x$\tau$ frames) from the full-length video and randomly crop 224\x224 pixels from a video, or its horizontal flip, with a shorter side randomly sampled in [256, 320] pixels.  

\paragraph{Inference.} 	Following common practice, in video classification~\cite{Feichtenhofer19}, we report 30-view, top-1 classification accuracy on the Kinetics validation set. We uniformly sample 10 clips from a video along its temporal axis. For each clip, we scale the shorter spatial side to 256 pixels and take 3 crops of 256\x256 to cover the spatial dimensions. We average the softmax scores for prediction.

\subsection{Details: AVA Action Detection}\label{sec:detection}

\paragraph{Dataset.}
The AVA dataset \cite{Gu2018} has bounding box annotations for spatiotemporal localization of (possibly multiple) human actions. It has 211k training and 57k validation video segments. We follow the standard protocol reporting mean Average Precision (mAP) on 60 classes \cite{Gu2018} on AVA v2.2.

\paragraph{Detection architecture.}
We exactly follow the detection architecture in \cite{Feichtenhofer19} to allow direct comparison of the pre-trained models used as a backbone for the AVA task~ \cite{Gu2018}.
The detector is similar to Faster R-CNN \cite{Ren16} with minimal modifications adapted for video. 
Region-of-interest (RoI) features \cite{Girshick15} are extracted at the last feature map of res$_5$ (\textit{cf.}~\tblref{tab:arch}) by extending a 2D proposal at a frame into a 3D RoI by replicating it along the temporal axis, followed by application of frame-wise RoIAlign \cite{He17}  and temporal global average pooling.
We set the spatial stride of res$_5$ to 1 (instead of 2), and use a dilation of 2 for its filters~\cite{Feichtenhofer19}. This increases the spatial resolution of res$_5$ by 2$\times$.
The RoI features are then max-pooled and fed to a per-class sigmoid classifier for prediction.

\paragraph{Training.} For direct comparison, the training procedure and hyper-parameters for AVA follow \cite{Feichtenhofer19} without modification. 
The network weights are initialized from the Kinetics models and we use step-wise learning rate decay, that is reduced by 10\x~after 16, 24 and 28 epochs. 
We train for 32 epochs on $\app$211k data, with linear warm-up \cite{Goyal17} for the first 5 epochs and use a weight decay of 10$^{-7}$, as in \cite{Feichtenhofer19}. 
For 8 GPU training, we use a batch-size of 64, a learning rate of 0.05 for the supervised pre-trained Kinetics models and 0.3 for the unsupervised ones, as this gives the best result for each of them.

The region proposal extraction also follows \cite{Feichtenhofer19} and is summarized here for completeness. 
Our region proposals are computed by an off-the-shelf person detector, \ie, that is not jointly trained with the action detection models.
We adopt a person-detection model trained with \emph{Detectron} \cite{Detectron2018}. It is a Faster R-CNN with a ResNeXt-101-FPN backbone.
It is pre-trained on ImageNet and the COCO human keypoint images~\cite{Lin2014}.
We fine-tune this detector on AVA for person (actor) detection.
The person detector produces 93.9 AP@50 on the AVA validation set.
Then, the region proposals for action detection are detected person boxes with a confidence of $>$ 0.8, which has a recall of 91.1\% and a precision of 90.7\% for the person class.

\paragraph{Inference.} We perform inference on a single clip with 8~frames sampled with stride 8~centered at the frame that is to be evaluated.

\subsection{Details: Charades Action Classification}\label{sec:charades}
\paragraph{Dataset.}
Charades \cite{Sigurdsson2016} has $\app$9.8k training videos and 1.8k validation videos in 157 classes in a multi-label classification setting of longer activities spanning $\app$30 seconds on average. Performance is measured in mean Average Precision (mAP).

\paragraph{Training.}
For {Charades}, we fine-tune the Kinetics models, but extend their duration by~2\x~($T$\x $\tau$ = 16\x 8) to account for the long-term nature of the dataset. This increase accuracy of all models by $\app$3 mAP. 
Our training augmentation is the same as as in \S\ref{sec:kinetics}. 
A per-class sigmoid output is used for mutli-class prediction. We train for 60 epochs using a batch size of 64 and a base learning rate of 0.2 (for 8 GPUs) with 10\x~step-wise decay at epoch 40 and 50, after warm-up in the first 5 epochs. We use weight decay of 10$^\text{-4}$ and dropout of 0.5.  Other training details are analogous to Kinetics.
\paragraph{Inference.} 
This is as for Kinetics (\S\ref{sec:kinetics}), but to infer the actions over a single video, we spatiotemporally max-pool prediction scores in testing \cite{Feichtenhofer19}.

\subsection{Details: Something-Something V2 (SSv2)}\label{sec:ssv2}

\paragraph{Dataset.}
The Something-Something V2 dataset~\cite{ssv2} contains 169k training, and 25k validation videos.
The videos show human-object interactions to be classified into 174 classes.
We report top-1 accuracy on the validation set.

\paragraph{Training.}
We fine-tune the pre-trained Kinetics models. We train for 22 epochs using a batch size of 64 and a base learning rate of 0.12 (for 8 GPUs) with 10\x~step-wise decay at epoch 14 and 18. Weight decay is set to 10$^{-6}$ and dropout 0.5. Our training augmentation is the same as in \S\ref{sec:kinetics}, but as Something-Something V2 requires distinguishing between directions, we disable random flipping during training.
We use segment-based input frame sampling~\cite{lin2018temporal} that splits each video into segments, and from each of them, we sample one frame to form a clip.

\paragraph{Inference.} We perform single center clip testing to form predictions over a single video.

\subsection{Details: UCF-101 Action Classification}\label{sec:ucf}

\paragraph{Dataset.}
UCF101 \cite{Soomro12} has 13320 human action videos in 101 categories. 
Our ablations are performed on the first train/val split, and for the comparison to prior work we report the mean average accuracy over the three splits. 

\paragraph{Training.}
We fine-tune the pre-trained Kinetics models and use the same augmentation as for Kinetics. We train for 200 epochs using a batch size of 64 and a base learning rate of 0.025 (for 8 GPUs) with 10\x~step-wise decay at epoch 60, 120 and 180. Weight decay is set to 0 and dropout to 0.8. 

\paragraph{Inference.} We use the same procedure as in Kinetics~(\S\ref{sec:kinetics}). 

\subsection{Details: HMDB-51 Action Classification}\label{sec:hmdb}

\paragraph{Dataset.}
HMDB51 \cite{Kuehne11} contains 6766 videos that have been annotated for 51 actions. Our evaluation follows the protocol for UCF101.

\paragraph{Training and Inference.}
Our settings are \textit{identical} to the ones used for UCF101 and we expect further tuning of hyper-parameters to increase its downstream performance. 


{\small
	\bibliographystyle{ieee_fullname}
	\bibliography{bib/shortstrings,bib/vgg_local,bib/vgg_other}
}

\end{document}